\pdfoutput=1

\documentclass[11pt]{article}

 \usepackage[final]{acl}
 
\usepackage{times}
\usepackage{latexsym}

\usepackage[T1]{fontenc}

\usepackage[utf8]{inputenc}

\usepackage{microtype}
\usepackage{amsmath} 
\usepackage{hyperref}
\usepackage{url}
\usepackage{graphicx}
\usepackage{multirow}
\usepackage{multicol}
\usepackage{adjustbox}
\usepackage{paralist}
\usepackage{booktabs}
\usepackage{makecell}
\usepackage{subfigure}
\usepackage{paralist}

\newlength{\mysize}
\newcommand{\mycfs}[1]{\setlength{\mysize}{#1pt}%
  \fontsize{\mysize}{1.2\mysize}\selectfont}

\usepackage{inconsolata}

\usepackage{graphicx}

\title{Learning with Less:  Knowledge Distillation from Large Language Models via Unlabeled Data}

\author{Juanhui Li$^1$\thanks{This work is done when Juanhui is the intern at Amazon},  Sreyashi Nag$^2$ ,  Hui Liu$^2$ , Xianfeng Tang$^2$,  Sheikh Sarwar$^2$, 
\\ {\bf Limeng Cui$^2$},
{\bf Hansu Gu$^2$}, {\bf Suhang Wang$^3$}, {\bf Qi He$^2$}, {\bf Jiliang Tang$^1$}\\ 
$^{1}$Michigan State University, 
$^{2}$Amazon.com,
$^{3}$The Pennsylvania State University\\
\texttt{\{lijuanh1,tangjili\}@msu.edu, szw494@psu.edu}\\
\texttt{ \{sreyanag,liunhu,xianft, smsarwar,culimeng,hansgu \}@amazon.com}
}

\begin{document}
\maketitle
\begin{abstract}

In real-world NLP applications, Large Language Models (LLMs) offer promising solutions due to their extensive training on vast datasets. However,  the large size and high computation demands of LLMs limit their practicality in
many applications, especially when further fine-tuning is required.
To address these limitations, smaller models are typically preferred for deployment. However, their training is hindered by the scarcity of labeled data. In contrast,  unlabeled data is often readily  which can be leveraged by using LLMs to generate pseudo-labels for training smaller models. This enables the smaller models (student) to acquire knowledge from LLMs (teacher) while reducing computational costs.
This process introduces challenges, such as   potential  noisy pseudo-labels.
Selecting high-quality and informative data is therefore critical to enhance model performance while improving the efficiency of data utilization.
To address this, we propose LLKD that enables Learning with Less computational resources and less data for Knowledge Distillation from LLMs. LLKD is
an adaptive sample selection method that incorporates signals from both the teacher and  student.  
Specifically, it prioritizes samples where the teacher demonstrates high confidence in its labeling, indicating reliable labels, and where the student exhibits a high information need, identifying challenging samples that require further learning. Our comprehensive experiments show that LLKD achieves superior performance across various datasets with higher data efficiency.
\end{abstract}

\section{Introduction}

Large Language Models (LLMs) such as LLaMA~\cite{touvron2023llama} and GPT-4~\cite{achiam2023gpt} have demonstrated superior language understanding abilities in many real-world NLP applications~\cite{schopf2022evaluating, thirunavukarasu2023large, zhao2023survey} due to the vast knowledge acquired from pre-training on extensive corpora.  However, deploying LLMs is resource-intensive with high memory requirements, computational costs, and increased latency during inference, especially when additional fine-tuning is needed for specific tasks~\cite{shoeybi2019megatron}.

To tackle these limitations, smaller models ~\cite{liu2019roberta,devlin2018bert, wang2024comprehensive} are often preferred due to their lower resource demands. Nonetheless, smaller models are not as powerful as LLMs~\cite{kaplan2020scaling} and typically require further training for specific tasks using labeled data, as they usually do not have the capacity to capture broad knowledge. Without the guidance of labeled data, self-supervised training on smaller models may lead to suboptimal performance, as these models struggle to generalize across diverse tasks and often fail to learn task-specific features effectively~\cite{goyal2019scaling}. This challenge is  further hindered by the high cost of obtaining task-related labeled data. While unlabeled data is generally more abundant, 
it cannot be directly utilized without proper labeling, posing a significant challenge for model training.

One promising approach is to use LLMs to generate pseudo-labels for unlabeled data, which can then be used to train smaller models. This strategy allows smaller models to benefit from the extensive knowledge embedded in the LLM while reducing computational costs. This process can be seen as a form of knowledge distillation~\cite{mishra2021confidence,zhou2023adads, kontonis2024slam, iliopoulos2022weighted}.
However, this approach presents challenges. Pseudo-labels generated by LLMs may be noisy or unreliable, potentially degrading the performance of the student model. 
Thus, achieving data efficiency is crucial—not only to reduce the impact of noisy pseudo-labels but also to ensure that representative data samples are selected for optimal training.

A  potential solution is to select data that not only has high pseudo-label quality but is also informative for the student  model.  However, as the student model continuously updates during training, identifying informative knowledge throughout this process remains a challenge.
Several existing works have proposed methods for data selection in the knowledge distillation process~\cite{mishra2021confidence, zhou2023adads, li2021dynamic}. However, most of these approaches~\cite{zhou2023adads, li2021dynamic} rely on datasets with true labels and do not consider the challenge of noisy pseudo-labeled samples, which can lead to suboptimal performance. While some methods~\cite{kontonis2024slam, iliopoulos2022weighted} address unlabeled data, they often overlook the student model’s learning progress or fail to consider data efficiency.
Therefore, it is beneficial to develop a method that
enables the student model to learn from the most valuable data while improving the data efficiency by reducing the amount of training data required.

To address these challenges, we propose  LLKD that enables Learning with Less computational resources and less data for Knowledge Distillation from LLMs. It is
an adaptive sample selection method for each training step that considers the student's dynamic learning status. We prioritize samples where the teacher model exhibits high confidence in its labeling, indicating reliable pseudo-labels~\cite{mishra2021confidence}, and the student model shows high uncertainty, pointing to challenging examples that require further learning~\cite{zhou2023adads}.
Specifically, we design two types of thresholds at each training step based on teacher confidence and student uncertainty, selecting overlapping samples that meet these criteria from both models’ perspectives. This data selection strategy promotes efficient knowledge transfer from the LLM to the smaller model, ensuring that the most informative samples are used for training while reducing the amount of data needed, thereby improving data efficiency.
We apply LLKD to a fundamental NLP task, text classification, and present comprehensive evaluation across various datasets. The results demonstrate that LLKD significantly enhances model performance, achieving superior results with higher data efficiency.

Our contributions are summarized as follows: 1) We introduce a knowledge distillation approach that leverages unlabeled data while requiring fewer computational resources.
2) We propose a dynamic data selection method LLKD that identifies high-quality samples, improving data efficiency.
3) Extensive experiments demonstrate that LLKD achieves superior text classification performance and enhanced data efficiency.

    


\section{Related Work}

\noindent \textbf{Knowledge Distillation}. 
Knowledge distillation~\cite{mishra2021confidence, zhou2023adads, xu2023computation, li2021dynamic, kontonis2024slam} has been used to transfer knowledge from a cumbersome teacher model to a lightweight student model. Most traditional methods focus on datasets with true labels, and some have explored data selection during this process. For example, \citet{mishra2021confidence} proposes a threshold based on learning epochs to select hard samples for student. Similarly, \citet{li2021dynamic} and \citet{xu2023computation} select samples with high student uncertainty by setting a fixed sampling ratio. \citet{zhou2023adads} introduces a reinforcement learning-based selector to measure student uncertainty in different ways.
However, most of these methods rely on true labels and do not address the issue of noisy pseudo-labels from the teacher model. While some approaches focus on unlabeled data~\cite{lang2022training, dehghani2018fidelity}, they often overlook data efficiency or fail to account for the student's evolving nature to identify informative samples for student. For instance, \citet{kontonis2024slam} generates new soft labels by combining the student’s soft labels with denoised teacher labels, and \citet{iliopoulos2022weighted} reweights the student's loss function to simulate loss with noise-free pseudo-labels.

\noindent \textbf{Thresholding Methods}.
In classification tasks with large amounts of unlabeled and noisy data, various confidence-based thresholding methods ~\citep{zhang2021flexmatch, sohn2020fixmatch, freematch, softmatch} have been proposed to prioritize samples with high confidence.
For example, 
FlexMatch~\citep{zhang2021flexmatch} employs a curriculum learning approach to flexibly adjust the threshold for each class based on the number of learned samples. FreeMatch~\cite{freematch} and SoftMatch~\cite{softmatch} use confidence-based thresholds, with FreeMatch considering both global and class-wise learning status, while SoftMatch weights the loss function using a Gaussian function. However, these methods often rely on limited labeled data, which can result in suboptimal performance with the self-training manner.



\begin{figure*}[t]
\begin{center}
 \centering
\includegraphics[width=0.95\linewidth]{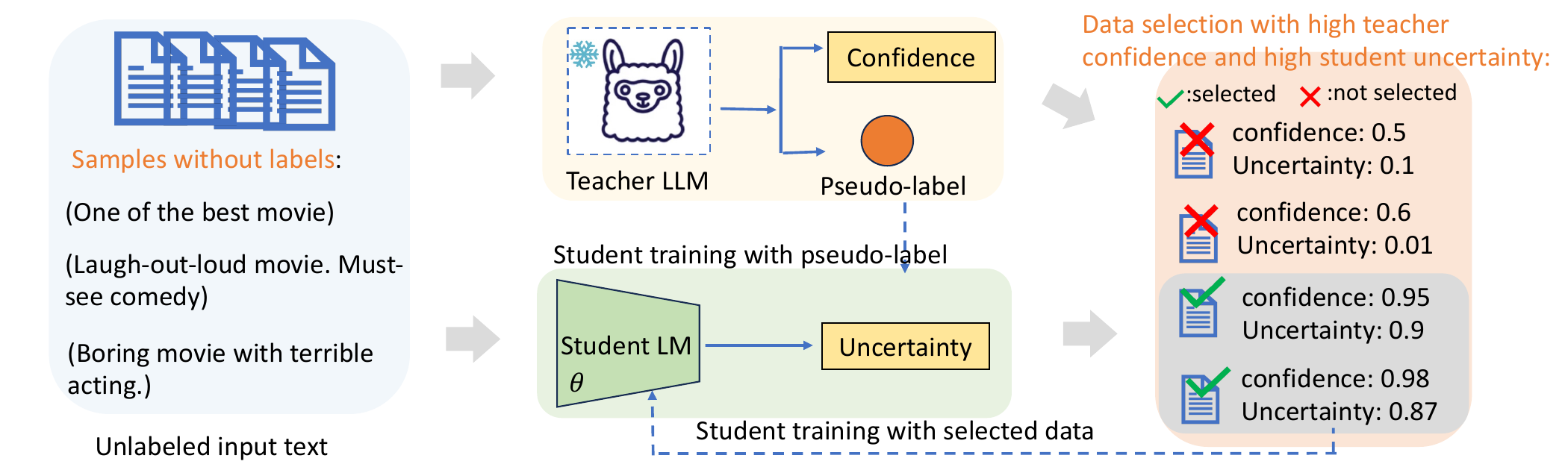} 
\caption{An illustration of the LLKD framework.  } 
\label{fig:framework}
\end{center}
\vspace{-0.2in}
\end{figure*}

\noindent \textbf{Unsupervised Text Classification}. It aims to categorize text without labeled data. A common approach is similarity-based methods~\cite{abdullahi2024retrieval, schopf2022evaluating, yin2019benchmarking}, which generate embeddings for both input texts and labels, and then match texts to labels based on similarity. These methods require no training data or training process. For example, ~\citet{abdullahi2024retrieval} suggests augmenting input texts with Wikipedia data, while ~\citet{schopf2022evaluating} uses lbl2TransformerVec to generate embeddings.
However, these methods often perform poorly without task-specific domain knowledge. Although some approaches~\cite{gretz2023zero} propose pre-training on datasets from other domains and applying the pretrained model for predictions, these models are often not publicly available due to legal and privacy concerns.

\section{Method}

\subsection{Notations}
Given the unlabeled dataset $\mathcal{X} =  \{x_1, x_2, ..., x_N\}$ where $ x_i \in \mathcal{X}$ is the input text, we do not have labels for the samples in $\mathcal{X}$. However, we are provided with a label set  $\mathcal{Y} = \{y_1, y_2, ..., y_K\}$  
 consisting of $K$ possible labels, where each $y_i \in \mathcal{Y}$ is the $i$-th label. For model training, we use $\mathcal{X}$ as the training set.
  To evaluate the performance, we have the validation and test set $\mathcal{X}_{dev}= \{ (x_j, y_j)\}_j^{M_1}$, $\mathcal{X}_{test}= \{ (x_j, y_j)\}_j^{M_2}$, where both sets contain the ground truth labels.  The validation set is used to select the model.

We leverage the unlabeled data to train a smaller model using pseudo-labels generated by the teacher LLM. Let $G_t$ denote the teacher LLM 
and $G_s$ represent the student language model. The framework of LLKD is illustrated in \figurename~\ref{fig:framework}.
It consists of three key components: 1) the teacher model, which generates pseudo-labels along with corresponding confidence scores. 
It remains fixed and is only used for inference; 
2)  the student model, which is trained on these pseudo-labels and generates uncertainty estimates for each sample; 
and 3) a data selection process, which selects samples based on both pseudo-label quality and informativeness for the student. Teacher confidence is used to assess pseudo-label quality, with higher confidence indicating more reliable labels. Meanwhile, the student's uncertainty is used to identify hard samples, where greater uncertainty suggests challenging knowledge that requires further training.
We integrate signals from both the teacher and student models to select samples that are both high in pseudo-label quality and informative for the student. These selected samples are then used to train the student model, significantly reducing the amount of data needed while enhancing data efficiency and model performance. Further details on each component are provided in the following subsections.


\subsection{Teacher Model}

Given the strong performance of LLMs in handling NLP tasks, we use LLaMA~\cite{touvron2023llama} as the teacher model to generate pseudo-labels. Typically, utilizing LLMs involves applying prompts with target inputs to guide the model in generating customized, task-specific outputs. These prompts are carefully designed templates that frame the input text for the LLM. They often include a few examples to illustrate the task, enabling the model to generalize from a limited amount of data by leveraging its extensive pre-training knowledge. This approach eliminates the need for additional fine-tuning, thereby enhancing the LLMs' flexibility and efficiency in real-world applications.
We select a few examples from the validation set to construct the prompt. More details are given in the section~\ref{sec:app_prompt} in the Appendix. Formally, we use $\mathcal{P}(x_i) $ to denote the prompt for input text $x_i$, then the pseudo-label is defined as follows:
\begin{equation}
    y(x_i)_{pl} = \arg\max_y  p_{G_t} (y | \mathcal{P}(x_i))
\end{equation}
where $ p_{G_t} (y | \mathcal{P}(x_i))$ is the probability vector  of the teacher over
the labels  given the input prompt $\mathcal{P}(x_i)$.
To measure the teacher's confidence in its prediction, we adopt a commonly used strategy~\cite{mishra2021confidence, zhou2023adads}, which utilizes the maximum probability assigned to the most probable label. This confidence measure is defined as: \begin{equation} C_i = \max_y p_{G_t} (y | \mathcal{P}(x_i)), \end{equation} where $C_i$ represents the maximum  teacher probability of sample $x_i$.

\subsection{Student Model}

For the student model, we utilize a smaller and more efficient pretrained language model, such as RoBERTa~\cite{liu2019roberta} denoted as $G_s$. It is observed that framing the classification task as a masked language modeling task within the prompt-learning framework yields superior performance compared to standard fine-tuning~\citep{yu2023cold, yin2019benchmarking}. This approach facilitates a clearer understanding of the task for the pretrained language model and enhances generalization~\citep{yin2019benchmarking}.
For example, a possible prompt template could be defined as $\mathcal{P}_s = [ x_i, \textit{it was [\text{MASK}]} ]$, where the $[\text{MASK}]$ token is used to predict the label. To measure the student model's uncertainty, we adopt a commonly used entropy-based approach, which relies on the probability distribution over the label set.
Using this template, we can derive the probability distribution over the label set $\mathcal{Y}$: 
\begin{align}
   \nonumber  p(y|x_i) &= p([\text{MASK}] = \mathcal{V}(y) | \mathcal{P}_s)\\
   &= \frac{\exp(\mathbf{h}_{\mathcal{V}(y)}^T \mathbf{h}_{[\text{MASK}]})}{\sum_{y'\in \mathcal{Y}}\exp(\mathbf{h}_{\mathcal{V}(y')}^T \mathbf{h}_{[\text{MASK}]})} 
\end{align}
Here, $\mathcal{V}(\cdot)$ is the verbalizer that maps each label $y$ to a word or words in the vocabulary of the student model, and $\mathbf{h}$ represents the token embedding generated by $G_s$. Specifically, $\mathbf{h}_{[\text{MASK}]}$  denotes the embedding of the $[\text{MASK}]$ token, while $\mathbf{h}_\mathcal{V}(y)$ corresponds to the embedding of the label word $\mathcal{V}(y)$.  As the student model is updated at each training step $t$, the uncertainty may vary at each step.  We formally define the uncertainty at each step as: 
\begin{equation}
    U_t(x_i) = - \sum_{j=1}^{K} p(y_j|x_i) \log p(y_j|x_i) 
\end{equation}
where $U_t(x_i)$ is the uncertainty of the student model on the sample $x_i$ at the $t$-the training step.


\begin{figure}[t]
\begin{center}
 \centerline{
{

\subfigure[]{
\includegraphics[width=0.52\linewidth]{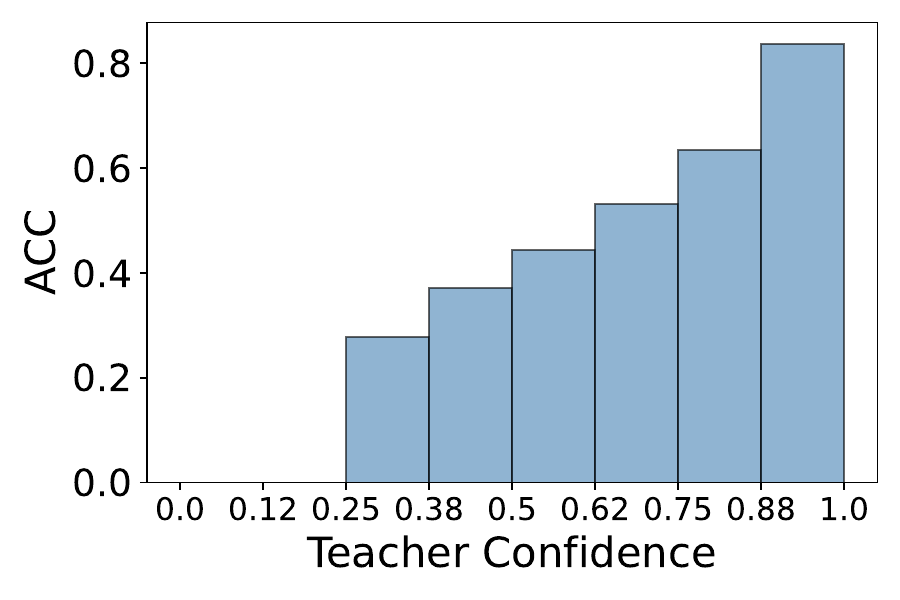} 
}
\subfigure[]{
\includegraphics[width=0.52\linewidth]{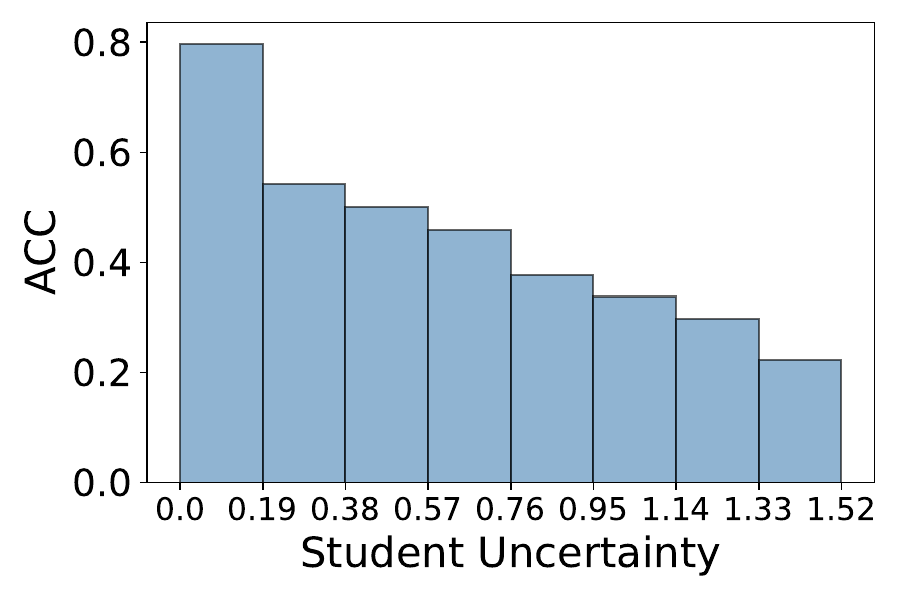} 
}
}
}
 
\caption{The relationship between teacher model accuracy and teacher confidence (a), and the relationship between student model accuracy and student uncertainty (b) on the validation set of Pubmed-RCT-20k dataset.    }
\label{fig:Acc_teacher_student}
\end{center}
\vspace{-0.3in}
\end{figure}

\subsection{Data Selection}

We propose leveraging teacher confidence and student uncertainty to identify samples with high-quality pseudo-labels and those that present challenging knowledge.
This is motivated by some empirical observations   showing that teacher confidence can assess pseudo-label quality, while student uncertainty indicates sample informativeness, as illustrated in \figurename~\ref{fig:Acc_teacher_student}. Additional results on more datasets are presented in Section~\ref{sec:app_emp} in the Appendix which show similar observations.
Based on the validation set of the PubMed-RCT-20k dataset (detailed in section~\ref{sec:dataset}), we generate pseudo-labels and compute teacher confidence with the teacher model. The confidence scores are grouped into bins, and the average accuracy is calculated for each bin, as shown in \figurename~\ref{fig:Acc_teacher_student}(a). Similarly, the student model provides predictions and student uncertainty, which are also binned to calculate average accuracy, as illustrated in \figurename~\ref{fig:Acc_teacher_student}(b). The student uncertainty varies across training steps, and the figure shows results from one step (we observe similar trends in other steps). We observe  that higher teacher confidence typically correlates with higher accuracy, signifying high-quality pseudo-labels, whereas higher student uncertainty generally corresponds to lower accuracy, reflecting incorrect predictions and harder samples.


To effectively leverage the two signals, we introduce two thresholds: one based on teacher confidence and the other on student uncertainty. As  the student model continuously updates during training and its learning status may differ across classes, necessitating an adaptive approach to define the thresholds as training progresses. Inspired from the FreeMatch framework~\cite{freematch}, which has shown strong performance in the image domain, we design the thresholds to incorporate both the global training status and the local class-specific status within each batch.
Although the teacher's confidence is not updated, we empirically found that determining the teacher confidence threshold on a batch-wise basis yields superior performance. Consequently, we adopt a similar batch-wise approach when setting the teacher confidence threshold. Formally, the global component of the student uncertainty threshold, which reflects the overall training status, is defined as follows:
\begin{equation}
    \tau_t^{S} =          \lambda_{S}\tau^{S}_{t-1} + (1-\lambda_{S}) \frac{1}{B} \sum_{i=1}^{B} U_{t}(x_i)
    \label{eq:global_S}
\end{equation}
where $t$ is the training step, $\tau_0^{S} = 0$ and $B$ is the batch size. $\lambda_{S} \in (0, 1) $ is the momentum in  Exponential Moving Average (EMA) over previous batches for a more stable estimation.
And the local part is computed based on each class $y$:
\begin{align}
  \nonumber  \hat{p}^{S}_t(y) =   & \lambda_{S}\hat{p}^{S}_{t-1}(y) + \\ &
    (1-\lambda_{S})* \frac{ \sum_{i=1}^{B}   \mathbf{1}(y(x_i)_{pl} =y) U_t(x_i)}{\sum_{i=1}^B \mathbf{1}(y(x_i)_{pl} =y)  }
    \label{eq:local_S}
\end{align}
where $\hat{p}^{S}_0(y) = 0$, $\mathbf{1}(\cdot)$ is the indicator function. It is 1 when $y(x_i)_{pl} =y$, and 0 otherwise. Then we obtain the final threshold based on the student uncertainty by integrating the global and local parts:
\begin{align}
   \nonumber \tau^{S}_t(y) &= \text{MaxNorm}(\hat{p}^{S}_t(y) )^{\beta_{S1}}* (\tau^{S}_t) ^{\beta_{S2}}
      \\ &= \bigg(\frac{\hat{p}^{S}_t(y)}{\max \{ \hat{p}^{S}_t(y) : y \in \mathcal{Y}\} }\bigg)^{\beta_{S1}}*(\tau^{S}_t)^{\beta_{S2}}
\end{align}
where MaxNorm is the Maximum Normalization, $\beta_{S1}$ and $\beta_{S2}$ are two hyper-parameters used to control the contribution of the local and global threshold. 
Similarly, by replacing the $U_t(x_i)$ as the teacher confidence $C_i$ in Eq.(\ref{eq:global_S}) and Eq.(\ref{eq:local_S}), we can obtain the final threshold $\tau^{T}_t(y)$ with its own momentum $\lambda_{T}$ and hyper-parameters $\beta_{T1}$ and $\beta_{T2}$. Then we combine $ \tau^{T}_t(y) $ and  $\tau^{S}_t(y)$ to select samples with high teacher confidence and high student uncertainty. The loss function for one batch at step $t$ is defined as follows:
\begin{align}
  \nonumber &   \mathcal{L} = 
 \frac{1}{B} \sum_{i=1}^B 
   \bigg (\mathbf{1}\big( U_t(x_i) \ge \tau^{S}_t(y(x_i)_{pl})\big)* \\ & \mathbf{1} \big(C_i \ge  \tau^{T}_t(y(x_i)_{pl})\big) 
    *\mathcal{H}\big( y(x_i)_{pl}, p(y|x_i) \big)\bigg )
    \label{eq:loss}
\end{align}
where $\mathcal{H}$ is the cross entropy loss. Furthermore, we apply a weighting scheme to the selected samples based on teacher confidence and student uncertainty. Specifically, we use a simple approach by adding the teacher confidence and student uncertainty to calculate the weight. The modified loss function is then defined as follows:
\begin{equation}
     \mathcal{L}_{w} = 
   \frac{1}{B} \sum_{i=1}^B(f(C_i)+f(U_t(x_i) ))*\mathcal{L}_i
   \label{eq:weight_loss}
\end{equation}
where $f$ is a sum normalization function across samples in the batch, $\mathcal{L}_i$ is the loss of each sample in the batch in Eq~(\ref{eq:loss}), i.e.,  $\mathcal{L} = 
 \frac{1}{B} \sum_i^B \mathcal{L}_i $.

\section{Experiment}
In this section, we conduct comprehensive experiments to validate the performance of LLKD. We will introduce the experimental settings, followed by results and their analysis.

\subsection{Experimental Settings}

\begin{table*}[!ht]
\centering
\caption{ ACC and Macro-F1  (\%) of the unsupervised text classification. Average results with standard deviation based on 3 seeds are reported. The bold and underline highlight the best and second-best results respectively.
}
\label{table:main_result}
\begin{adjustbox}{width =1 \textwidth}
\begin{tabular}{l|cc|cc|cc|cc|cc| c}
\toprule
Models &\multicolumn{2}{|c|}{PubMed-RCT-20k} & \multicolumn{2}{c|}{Yahoo! Answers} & \multicolumn{2}{c|}{Emotions} & \multicolumn{2}{c|}{Arxiv-10} &\multicolumn{2}{c|}{BiosBias}&  \makecell[c]{Avg.\\Rank} \\
\midrule
& ACC & F1 &ACC &F1 &ACC &F1 &ACC &F1 &ACC &F1&/\\
\midrule
Teacher-ZS & 53.67	& 44.77 & 57.84	&58.91	&55.15&	45.64	&53.45&	53.08	&68.76	&51.9 &11.9\\
Teacher &67.09	&57.88&	65.45	&64.12	&59.28&	48.6	&56.02&	56.5	&75.83	&61.32 &10.4\\
Random &  70.35 ± 0.73	&60.70 ± 1.23	&65.91 ± 0.31	&64.58 ± 0.42&	64.60 ± 1.40	&54.35 ± 0.68&	60.35 ± 1.56	& 60.19 ± 1.52	&76.11 ± 0.47&	61.18 ± 1.35&7.1\\
No\_DS &  68.99 ± 0.88	&60.40 ± 0.47	&65.38 ± 0.92	&64.06 ± 0.75	&64.55 ± 1.53	& 54.79 ± 2.34&	60.57 ± 0.54	&60.16 ± 0.53	&75.96 ± 0.46	& 62.09 ± 0.80&8.2\\
\midrule
FreeMatch& 69.17 ± 0.39	&60.31 ± 1.17	& 65.98 ± 0.41	&64.26 ± 0.12&	64.81 ± 0.34&	 52.68 ± 0.94	& 60.58 ± 0.91	& 60.36 ± 1.17&	75.63 ± 0.32&	 61.37 ± 0.76 &8.1\\
SoftMatch& 69.63 ± 0.97	&60.56 ± 1.51	& 66.47 ± 0.16&	64.70 ± 0.84	&65.48 ± 0.46	&54.64 ± 1.03	&61.01 ± 0.81	& 59.62 ± 0.78&	76.09 ± 0.35&	61.20 ± 1.20 &6.4\\
\midrule 
CCKD\_L&	69.43 ± 0.51	& 60.61 ± 1.11	& 65.93 ± 0.32	&64.27 ± 0.41	& 63.19 ± 1.31	 &52.70 ± 1.35	&61.15 ± 0.18&	60.10 ± 1.25	&76.41 ± 0.17	&61.84 ± 0.80&7.3 \\
CCKD\_T+Reg&	69.89 ± 0.69	&60.97 ± 0.87	&66.36 ± 0.51	&64.53 ± 0.05	&64.06 ± 1.32	&52.89 ± 0.59	&60.75 ± 1.62&	60.74 ± 1.51	&76.43 ± 0.12&	64.04 ± 1.01 &5.8\\
UNIXKD	&  70.47 ± 0.97	&61.65 ± 0.59	&66.56 ± 0.70	&64.49 ± 1.05	&65.29 ± 0.45	&\underline{55.64 ± 1.15}	& 61.37 ± 1.31	&61.35 ± 1.07	&75.94 ± 0.51&	61.86 ± 0.85 &4.6\\
Entropy Score& 71.16 ± 0.91	&62.53 ± 0.84	&66.61 ± 0.29	&65.08 ± 0.53	&64.62 ± 0.78	&55.42 ± 0.83	&60.86 ± 0.57&	59.84 ± 0.45	&76.62 ± 0.19&	64.18 ± 0.48 &4.2 \\
\midrule
\makecell[l]{Lbl2TransformerVec}&	31.82	&23.44	&53.74	&53.31	&42.73	&39.67&	42.88	&41.71	&47.78&	47.32 &12.9\\
\midrule
LLKD &  \underline{ 73.46 ± 0.20}	& \underline{65.72 ± 0.22}	& \underline{66.82 ± 0.49}	&\textbf{65.5 ± 0.06}	&\textbf{67.04 ± 1.12}	&55.42 ± 2.62	&\underline{62.90 ± 0.99}	& \underline{62.18 ± 1.08}	& \textbf{76.94 ± 0.13}&	\textbf{64.63 ± 0.30} &1.7\\
LLKD\_w & \textbf{74.07 ± 0.94} &	\textbf{66.17 ± 1.22}&	 \textbf{66.83 ± 0.12}	& \underline{65.48 ± 0.46}&	 \underline{67.03 ± 1.65}	& \textbf{56.24 ± 1.04}&	\textbf{62.99 ± 0.33}&	\textbf{ 62.90 ± 0.15}	&\underline{76.77 ± 0.16}&	\underline{64.46 ± 0.43} &1.4\\
Improv. & 4.09\%	&5.82\%	&0.33\%	&0.80\%	&2.38\%	&1.07\%	&2.64\%	&2.53\%	&0.42\%	&0.70\% &/\\
\bottomrule
\end{tabular}
\end{adjustbox}
\vspace{-0.1in}
\end{table*}


\label{sec:dataset}
\noindent \textbf{Datasets.}
We use five datasets from various domains: \textbf{PubMed-RCT-20k}~\citep{dernoncourt2017pubmed}, extracted from medical papers; \textbf{Yahoo! Answers}~\citep{zhang2015character}, a collection of question-answer pairs from the Yahoo! Answers platform; \textbf{Emotions}~\citep{saravia-etal-2018-carer}, which contains Twitter messages categorized into six basic emotions; \textbf{Arxiv-10}~\citep{farhangi2022protoformer}, built from ArXiv papers; and \textbf{BiosBias}~\citep{De-ArteagaRWCBC19}, a dataset of textual biographies aimed at predicting professional occupations. More details are given in Section~\ref{sec:app_data} in the appendix.


\noindent \textbf{Baselines.}
We compare our method against four groups of baseline approaches:
1) Thresholding methods, such as \textbf{FreeMatch}~\citep{freematch}, which uses an adaptive threshold to select samples with high student confidence, and \textbf{SoftMatch}~\citep{softmatch}, which assigns higher weights to samples with greater student confidence.
2) Knowledge distillation methods,  which focus on filtering noisy pseudo-labels based on the teacher or selecting informative samples based on the student. For the first type, we evaluate \textbf{CCKD\_L}~\cite{mishra2021confidence}, which weights samples according to the teacher's probability, and \textbf{Entropy Score}~\citep{lang2022training}, which selects samples with the lowest teacher entropy, indicating high teacher confidence. For the second type, we include \textbf{CCKD\_T+Reg}~\cite{mishra2021confidence}, which uses a threshold to select challenging samples for the student, and \textbf{UNIXKD}\citep{xu2023computation}, which selects samples with the highest student uncertainty.
3) Unsupervised text classification method:
\textbf{Lbl2TransformerVec}\citep{schopf2022evaluating}, which predicts labels based on the similarity between the embeddings of input text and label words. 4) Basic baselines: \textbf{Random}, which selects a random subset of samples from each batch; \textbf{Teacher}, which generates predictions directly using the teacher model with few-shot examples; \textbf{Teacher-ZS}, which generates predictions using the teacher model without examples (zero-shot); and \textbf{No\_DS}, where the student model is trained without data selection. 

\noindent \textbf{Implementation details.}
For the teacher model, we use LLaMA~\cite{touvron2023llama}, an open-source LLM that has demonstrated strong performance in various applications. For the student model, we adopt RoBERTa~\cite{liu2019roberta}. To ensure a fair comparison, all baseline models use the same pseudo-labels as our model, and the baseline models  use RoBERTa as the backbone model except for the Lbl2TransformerVec.
Model performance on the text classification task is evaluated using Accuracy (ACC) and Macro-F1 scores, the data efficiency is evaluated based on the total used training number. We provide a  parameter analysis in Section~\ref{sec:parameter_analysis}.  
Additional details are given in  Section~\ref{sec:app_prompt} and Section~\ref{sec:app_settings} in the Appendix.

\begin{table*}[!ht]
\centering
\caption{ 
Data efficiency: total number of training samples and their percentage of the original dataset. }
\label{table:data_effi}
\begin{adjustbox}{width =0.95 \textwidth}
\begin{tabular}{l|ccccc}
\toprule
Models & PubMed-RCT-20k &Yahoo! Answers & Emotions &Arxiv-10 &BiosBias \\
\midrule

FreeMatch &469,218 (87\%) &	266,314 (88.8\%)	&240,144 (72\%)&	415,222 (86.5\%)&	356,578 (92.4\%)\\
CCKD\_T+Reg& 214,967 (39.8\%)	&118,502 (39.5\%)	&242,133(72.6\%)&	449,526(93.9\%)&	195,307 (50.6\%)   \\
UNIXKD & 151,875 (28.1\%)	&262,500 (87.5\%)	&93,780 (28.1\%)	&\textbf{45,000 (9.4\%)}&	337,876 (87.5\%) \\
Entropy Score &270,045 (50\%)&	209,937	(70\%)&233,401 (70\%) &335,874 (70\%)	&347,481 (90\%)\\
\midrule
LLKD& \textbf{19,828 (3.7\%)}	& \textbf{101,396 (33.8\%)}	& \textbf{80,725 (24.2\%)}	& {161,008 (33.6\%)}	& \textbf{39,423 (10.2\%) }\\
w/o SU & 366,423 (67.8\%)	&238,107 (79.4\%)	&197,757 (59.3\%)	&376,080 (78.4\%)	&315,512 (81.7\%)\\
w/o TC&  71,662 (13.3\%)	&155,547 (51.9\%)	&150,091 (45\%)	&242,105 (50.5\%)	&83,259 (21.6\%)\\

\bottomrule
\end{tabular}
\end{adjustbox}
\vspace{-0.1in}
\end{table*}

\subsection{Classification Performance Comparison}
We present the classification performance in Table~\ref{table:main_result}, where 
``LLKD\_w'' denotes the version with weighting in the loss function, as defined in Eq~(\ref{eq:weight_loss}). 
We also report the average rank for each method over all datasets and metrics. The relative improvement of our best method
over the best baseline is indicated as ``Improv.''    
The  Lbl2TransformerVec is the similarity-based method without any training progress, thus it does not have standard deviation. 

Our key observations are as follows:
1) As shown in Table~\ref{table:main_result}, our model consistently outperforms all baseline methods, achieving a significant relative improvement of 5.82\% on the Pubmed-rct-20k dataset in terms of F1 score. The weighted version generally performs better, indicating that leveraging teacher confidence and student uncertainty to prioritize selected samples further enhances overall model performance.
2) Generally, we  observe that directly using the teacher model performs worse than our method, as well as all other baselines (except for {Lbl2TransformerVec}), which fine-tune the student model using pseudo-labels.
This demonstrates that the student model not only effectively learns from the teacher but also achieves superior results. These findings suggest that, with proper tuning, the student model can deliver better performance while maintaining much lower computational costs. 3) The teacher model (Teacher) with few-shot examples generally outperforms Teacher-ZS, demonstrating the effectiveness of incorporating few-shot examples.
4) Additionally, the  similarity-based method, Lbl2TransformerVec, exhibit the weakest performance, highlighting that relying solely on text-label similarity is insufficient for effective classification.



\subsection{Ablation Study}
\begin{figure}[t]
\begin{center}
 \centering{
\includegraphics[width=0.99\linewidth]{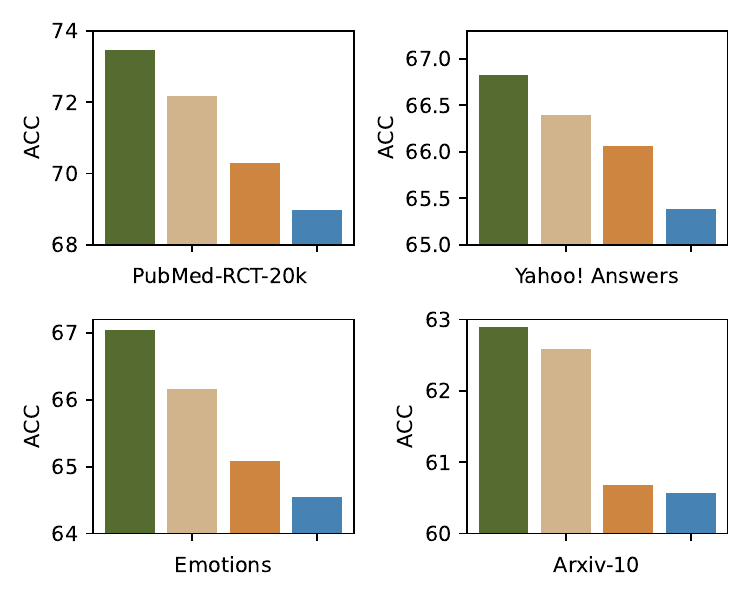} 
}
 \centering{
 \includegraphics[width=0.65\linewidth]{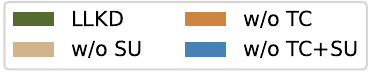} 
 }
\caption{Ablation study on various datasets.  }
\label{fig:ablation_study}
\end{center}
\vspace{-0.2in}
\end{figure}

In this subsection, we conduct an ablation study to evaluate the effectiveness of each component in our method, including the data selection, teacher confidence, and student uncertainty.  The results are shown in \figurename~\ref{fig:ablation_study}. We use ``w/o TC'' to denote the model without using the teacher confidence to select samples, relying solely on the student uncertainty threshold.   Similarly, ``w/o SU''  denotes the model without using the student uncertainty threshold to select, while ``w/o TC+SU'' represents the model without any data selection. Note that ``w/o TC+SU'' is identical to the No\_DS in Table~\ref{table:main_result}. The figure clearly shows that our model consistently outperforms the ablated versions across all datasets, highlighting the importance of each component. Notably, when no data selection is performed, the model exhibits the worst performance, further validating the critical role of our data selection strategy in enhancing the model performance.

Additionally, to assess the necessity of utilizing unlabeled data when labeled data is limited, we conduct an analysis, with further details provided in Section~\ref{sec:app_true_label} in the Appendix. The results in Table~\ref{table:true_label} indicate that training with a limited number of true labels yields significantly worse performance compared to the LLKD model trained with unlabeled data. This finding highlights the importance of leveraging unlabeled data when labeled data is limited.

\subsection{Data Efficiency}
In this subsection, we evaluate data efficiency by presenting the total number of  samples seen and the  percentage of original total  samples seen in each run. For example, the original training size of Arxiv-10 is 79,790 and we set the epoch to be 6,  so the student model sees a total of  79,970 x 6 = 479,820 samples without any data selection.  All methods have the same original total seen samples. 
The results of various data selection method are shown in Table~\ref{table:data_effi}. 
We also include results from our ablated versions to provide deeper insights into our approach. For UNIXKD and Entropy Score which have fixed selection ratio, we experimented with  ratios of \{10\%, 30\%, 50\%, 70\%, 90\%\} and used the ratio that achieved the best validation performance. 
Since SoftMatch and CCKD\_L weight samples instead of selecting, they use the full set of original   samples. 
Generally, 
the results indicate that our method consistently outperforms others in both effectiveness and data efficiency. In most cases, our approach requires selecting less than 25\% of the training samples. Notably, on the PubMed-RCT-20k dataset, we use as little as 3.7\% of the training samples while achieving a significant relative improvement of 5.82\%, as shown in Table~\ref{table:main_result}. Although the UNIXKD is trained with less data on the Arxiv-10 dataset, our method has large performance improvement, indicating UNIXKD does not select sufficient informative data.


\subsection{Choice of Teacher Model}
To evaluate if LLKD is agnostic of the choice of teacher LLM,  we try Gemma\footnote{\url{https://huggingface.co/google/gemma-2-9b-it}} \cite{team2024gemma} as the teacher model which  has  recently demonstrated strong performance in various applications.  The results on the Arxiv-10 dataset are presented in Table~\ref{table:gemma}.  
\begin{table}[!t]
\centering
\caption{  Comparison results (\%) by adopting Gemma as the teacher model on the Arxiv-10 dataset.}
\label{table:gemma}
\begin{adjustbox}{width =0.45 \textwidth}
\begin{tabular}{l|cc}
\toprule
Models & ACC & F1\\
\midrule
Teacher&  59.5	&59.5\\
Random & 61.56 ± 0.85	&60.86 ± 1.03 \\
No\_DS & 60.87 ± 0.66	&59.72 ± 0.67\\
FreeMatch & 60.98 ± 0.79	& 60.30 ± 1.35\\
SoftMatch & 61.55 ± 1.06	&60.96 ± 0.91\\
CCKD\_L & 60.68 ± 0.46	& 59.98 ± 0.65\\
CCKD\_T+Reg & 61.17 ± 0.64	&59.44 ± 1.82\\
UNIXKD &  61.40 ± 0.60	&61.08 ± 0.83 \\
Entropy Score & 60.34 ± 0.95&	59.53 ± 1.17\\
Lbl2TransformerVec & 42.88	& 41.71 \\
LLKD & \textbf{62.84 ± 0.68} &	\textbf{62.09 ± 0.71}\\
\bottomrule
\end{tabular}
\end{adjustbox}
\end{table}
We observe that Gemma outperforms LLaMA, achieving 59.5\% accuracy compared to LLaMA's 56.02\%. The baseline models also show slightly improved performance when using Gemma instead of LLaMA.
LLKD continues to demonstrate the best performance across all scenarios. Since Lbl2TransformerVec relies on text-label similarity, its performance remains unchanged regardless of the teacher model. Notably, these results indicate that LLKD is agnostic to the choice of teacher model and can be effectively applied with various LLMs.

\begin{figure}[t]
\begin{center}
 \centerline{
{


}
}
 \centerline{
\subfigure{
\includegraphics[width=1\linewidth]{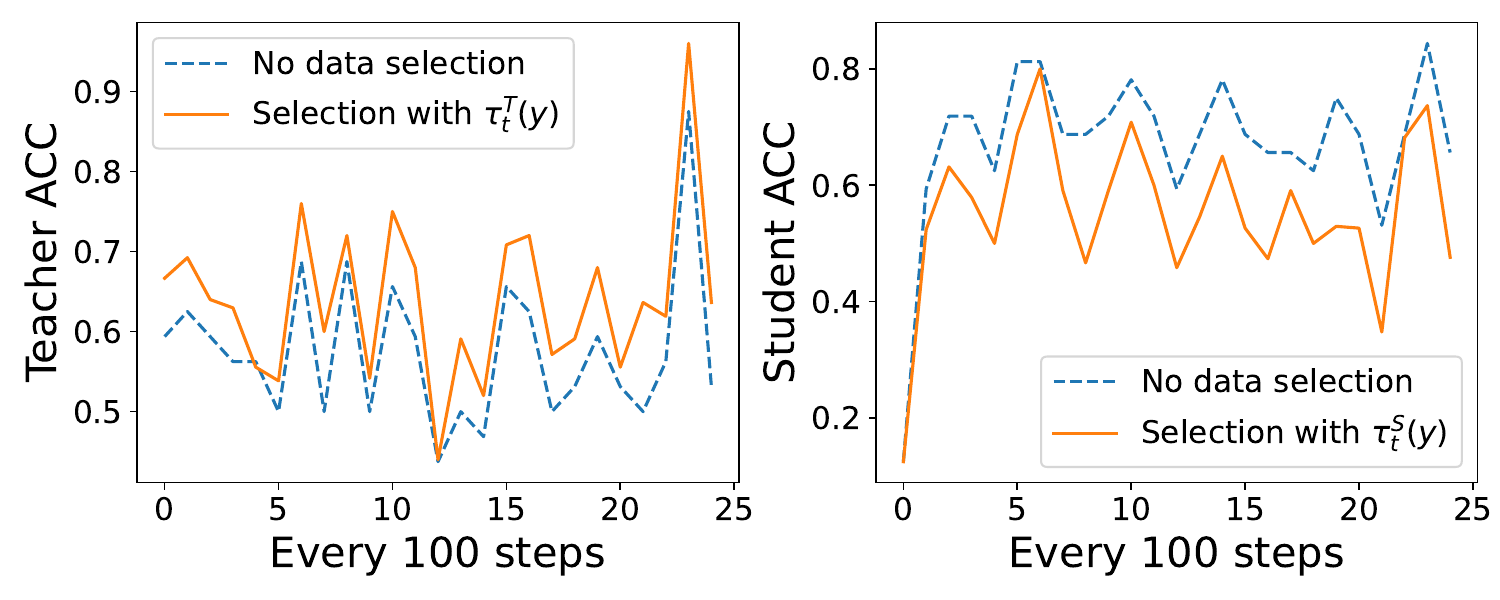} 
}
 }

\caption{Teacher ACC and student ACC before and after data selection on the Arxiv-10 dataset.  }
\label{fig:threshold}
\end{center}
\vspace{-0.2in}
\end{figure}

\subsection{Thresholds Evaluation}
In this subsection, we evaluate whether the thresholds $ \tau^{T}_t(y) $ (based on teacher confidence) and $ \tau^{S}_t(y) $ (based on student uncertainty) effectively select high-quality pseudo-labeled samples and hard samples from the training set, respectively. Specifically, we compute teacher accuracy by comparing the teacher’s pseudo-label with the true label in the training set (note that the true label is only used for evaluation here and is never involved in our method). Student accuracy is calculated by comparing the student’s predictions with the pseudo-labels. Teacher accuracy reflects the quality of pseudo-labels, with higher accuracy indicating a greater likelihood that the pseudo-label is correct. Student accuracy, on the other hand, identifies difficult samples, where lower accuracy suggests a higher likelihood of incorrect predictions, marking these as hard samples.
The results, shown in \figurename~\ref{fig:threshold}, are presented every 100 training steps for the ArXiv-10 datasets. Additional results on other datasets are given in Section~\ref{sec:app_threshold} in the appendix, which show similar observations. 
\figurename~\ref{fig:threshold} reveals that using $ \tau^{T}_t(y) $ tends to yield higher teacher accuracy, while selecting samples with $ \tau^{S}_t(y) $ leads to lower student accuracy, indicating the selection of harder samples. It demonstrates that the two thresholds work as expected.


\subsection{Hyper-parameter Analysis}
\begin{figure}[t]
\begin{center}
 \centerline{
{
\subfigure[]{
\includegraphics[width=0.48\linewidth]{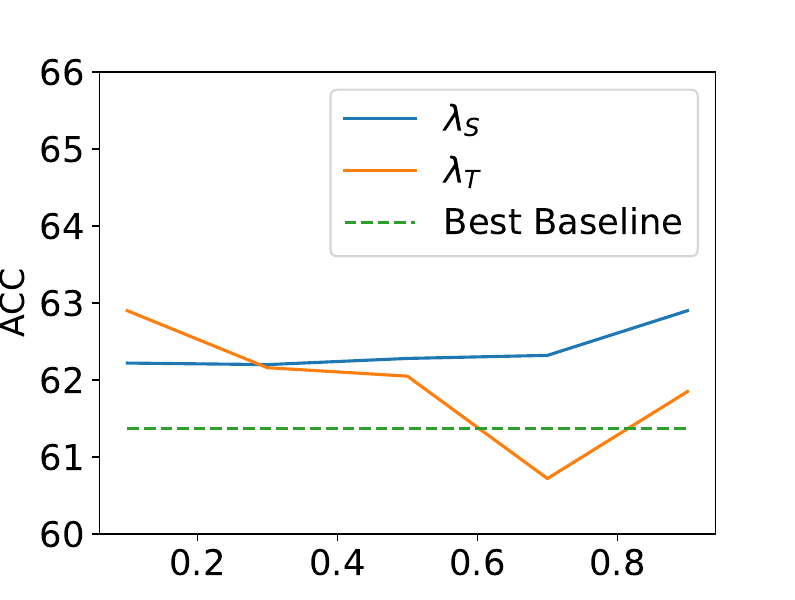} 
}
\subfigure[]{
\includegraphics[width=0.48\linewidth]{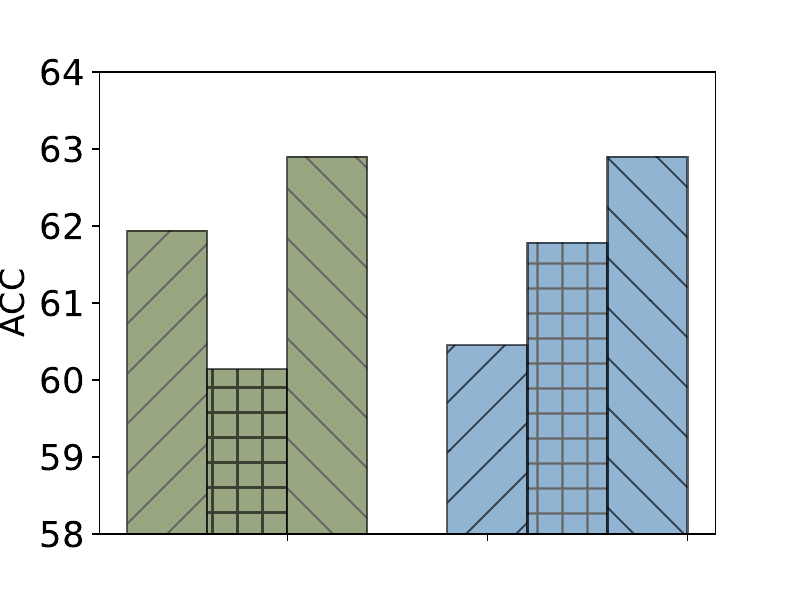} 
}
}
}


\centerline{
{\subfigure{
\includegraphics[width=1\linewidth]{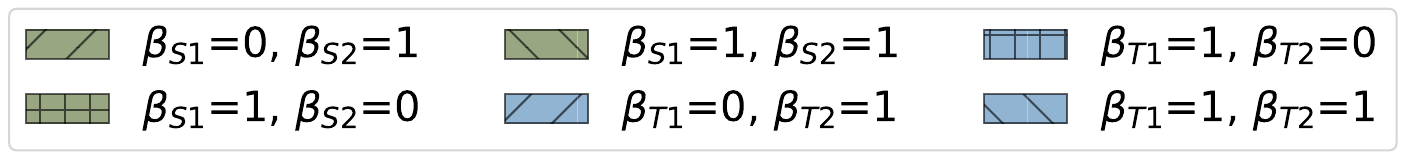} 
}}
}
 
\caption{Parameter analysis on the  Arxiv-10 dataset.  }
\label{fig:parameter_analysis}
\end{center}
\vspace{-0.4in}
\end{figure}

We analyze the sensitivity of key hyper-parameters used to define thresholds for teacher confidence and student uncertainty: $\lambda_S$ and $\lambda_T$, which control the momentum between previous and current learning status, and $\beta_{S1}, \beta_{S2}, \beta_{T1}$, and $\beta_{T2}$, which balance the contributions of global and local components in $\tau^{S}_t(y)$ and $\tau^{T}_t(y)$.
The results on the Arxiv-10 dataset are shown in \figurename~\ref{fig:parameter_analysis}. We vary $\lambda_S$ and $\lambda_T$ within the range \{0.1, 0.3, 0.5, 0.7, 0.9\} and observe that our model remained robust, consistently outperforming the best baseline in almost all cases. For $\beta_{S1}, \beta_{S2}, \beta_{T1}$, and $\beta_{T2}$, we test values in \{0, 1\} to assess the effects of using global, local, or both components. Specifically, when $\beta_{S1}=1$ and $\beta_{S2}=0$, the threshold involves only the local part; when $\beta_{S1}=0$ and $\beta_{S2}=1$, it involves only the global part; and when both $\beta_{S1}=1$ and $\beta_{S2}=1$, it considers both. The same applies to $\beta_{T1}$ and $\beta_{T2}$.
Our observations indicate that the best performance is achieved when both global and local components are considered, demonstrating their combined effectiveness in enhancing model performance.

\label{sec:parameter_analysis}
\section{Conclusion}
In this work, we propose to learn with less computational resources and less data for knowledge distillation from LLMs via unlabeled data.
The student model is fine-tuned with pseudo-labels generated by the teacher LLM. 
We empirically demonstrate that teacher confidence can effectively indicate the quality of pseudo-labels, while student uncertainty signals whether a sample contains challenging knowledge for the student. Specifically, we propose an adaptive data selection method to select the most informative samples, improving data efficiency by reducing the required training data. We introduce two thresholds based on teacher confidence and student uncertainty. By selecting samples that meet both thresholds, we ensure the chosen data has high pseudo-label quality and contains challenging knowledge. The effectiveness of the proposed method is validated through extensive experiments on various datasets, showing superior model performance and higher data efficiency.

\section{Limitations}
The experiments in this paper are limited to the text classification setting and do not explore other tasks and architectures commonly used in knowledge distillation, such as generation tasks. Additionally, due to resource constraints, we were unable to perform a thorough  study on different LLM sizes, which could be important to further demonstrate the generalizability of our approach.

\section{Acknowledgements}
This research is supported by the National Science Foundation (NSF) under grant numbers CNS2321416, IIS2212032, IIS2212144, IOS2107215, DUE2234015, CNS2246050, DRL2405483 and IOS2035472, the Army Research Office (ARO) under grant number W911NF-21-1-0198, Amazon Faculty Award, JP Morgan Faculty Award, Meta, Microsoft and SNAP.

\bibliography{custom}

\begin{thebibliography}{33}
\providecommand{\natexlab}[1]{#1}

\bibitem[{Abdullahi et~al.(2024)Abdullahi, Singh, and Eickhoff}]{abdullahi2024retrieval}
Tassallah Abdullahi, Ritambhara Singh, and Carsten Eickhoff. 2024.
\newblock Retrieval augmented zero-shot text classification.
\newblock In \emph{Proceedings of the 2024 ACM SIGIR International Conference on Theory of Information Retrieval}, pages 195--203.

\bibitem[{Achiam et~al.(2023)Achiam, Adler, Agarwal, Ahmad, Akkaya, Aleman, Almeida, Altenschmidt, Altman, Anadkat et~al.}]{achiam2023gpt}
Josh Achiam, Steven Adler, Sandhini Agarwal, Lama Ahmad, Ilge Akkaya, Florencia~Leoni Aleman, Diogo Almeida, Janko Altenschmidt, Sam Altman, Shyamal Anadkat, et~al. 2023.
\newblock Gpt-4 technical report.
\newblock \emph{arXiv preprint arXiv:2303.08774}.

\bibitem[{Chen et~al.(2023)Chen, Tao, Fan, Wang, Savvides, Wang, Raj, Xie, and Schiele}]{softmatch}
H~Chen, R~Tao, Yue Fan, Y~Wang, M~Savvides, J~Wang, B~Raj, X~Xie, and Bernt Schiele. 2023.
\newblock Softmatch: Addressing the quantity-quality tradeoff in semi-supervised learning.
\newblock In \emph{Eleventh International Conference on Learning Representations}. OpenReview. net.

\bibitem[{De{-}Arteaga et~al.(2019)De{-}Arteaga, Romanov, Wallach, Chayes, Borgs, Chouldechova, Geyik, Kenthapadi, and Kalai}]{De-ArteagaRWCBC19}
Maria De{-}Arteaga, Alexey Romanov, Hanna~M. Wallach, Jennifer~T. Chayes, Christian Borgs, Alexandra Chouldechova, Sahin~Cem Geyik, Krishnaram Kenthapadi, and Adam~Tauman Kalai. 2019.
\newblock Bias in bios: {A} case study of semantic representation bias in a high-stakes setting.
\newblock In \emph{Proceedings of the Conference on Fairness, Accountability, and Transparency, FAT* 2019, Atlanta, GA, USA, January 29-31, 2019}, pages 120--128. {ACM}.

\bibitem[{Dehghani et~al.(2018)Dehghani, Mehrjou, Gouws, Kamps, and Sch{\"o}lkopf}]{dehghani2018fidelity}
M~Dehghani, A~Mehrjou, S~Gouws, J~Kamps, and B~Sch{\"o}lkopf. 2018.
\newblock Fidelity-weighted learning.
\newblock In \emph{6th International Conference on Learning Representations (ICLR 2018)}. OpenReview. net.

\bibitem[{Dernoncourt and Lee(2017)}]{dernoncourt2017pubmed}
Franck Dernoncourt and Ji~Young Lee. 2017.
\newblock Pubmed 200k rct: a dataset for sequential sentence classification in medical abstracts.
\newblock In \emph{Proceedings of the Eighth International Joint Conference on Natural Language Processing (Volume 2: Short Papers)}, pages 308--313.

\bibitem[{Devlin(2018)}]{devlin2018bert}
Jacob Devlin. 2018.
\newblock Bert: Pre-training of deep bidirectional transformers for language understanding.
\newblock \emph{arXiv preprint arXiv:1810.04805}.

\bibitem[{Farhangi et~al.(2022)Farhangi, Sui, Hua, Bai, Huang, and Guo}]{farhangi2022protoformer}
Ashkan Farhangi, Ning Sui, Nan Hua, Haiyan Bai, Arthur Huang, and Zhishan Guo. 2022.
\newblock Protoformer: Embedding prototypes for transformers.
\newblock In \emph{Advances in Knowledge Discovery and Data Mining: 26th Pacific-Asia Conference, PAKDD 2022, Chengdu, China, May 16--19, 2022, Proceedings, Part I}, pages 447--458.

\bibitem[{Goyal et~al.(2019)Goyal, Mahajan, Gupta, and Misra}]{goyal2019scaling}
Priya Goyal, Dhruv Mahajan, Abhinav Gupta, and Ishan Misra. 2019.
\newblock Scaling and benchmarking self-supervised visual representation learning.
\newblock In \emph{Proceedings of the ieee/cvf International Conference on computer vision}, pages 6391--6400.

\bibitem[{Gretz et~al.(2023)Gretz, Halfon, Shnayderman, Toledo-Ronen, Spector, Dankin, Katsis, Arviv, Katz, Slonim et~al.}]{gretz2023zero}
Shai Gretz, Alon Halfon, Ilya Shnayderman, Orith Toledo-Ronen, Artem Spector, Lena Dankin, Yannis Katsis, Ofir Arviv, Yoav Katz, Noam Slonim, et~al. 2023.
\newblock Zero-shot topical text classification with llms-an experimental study.
\newblock In \emph{Findings of the Association for Computational Linguistics: EMNLP 2023}, pages 9647--9676.

\bibitem[{Iliopoulos et~al.(2022)Iliopoulos, Kontonis, Baykal, Menghani, Trinh, and Vee}]{iliopoulos2022weighted}
Fotis Iliopoulos, Vasilis Kontonis, Cenk Baykal, Gaurav Menghani, Khoa Trinh, and Erik Vee. 2022.
\newblock Weighted distillation with unlabeled examples.
\newblock \emph{Advances in Neural Information Processing Systems}, 35:7024--7037.

\bibitem[{Kaplan et~al.(2020)Kaplan, McCandlish, Henighan, Brown, Chess, Child, Gray, Radford, Wu, and Amodei}]{kaplan2020scaling}
Jared Kaplan, Sam McCandlish, Tom Henighan, Tom~B Brown, Benjamin Chess, Rewon Child, Scott Gray, Alec Radford, Jeffrey Wu, and Dario Amodei. 2020.
\newblock Scaling laws for neural language models.
\newblock \emph{arXiv preprint arXiv:2001.08361}.

\bibitem[{Kontonis et~al.(2024)Kontonis, Iliopoulos, Trinh, Baykal, Menghani, and Vee}]{kontonis2024slam}
Vasilis Kontonis, Fotis Iliopoulos, Khoa Trinh, Cenk Baykal, Gaurav Menghani, and Erik Vee. 2024.
\newblock Slam: Student-label mixing for distillation with unlabeled examples.
\newblock \emph{Advances in Neural Information Processing Systems}, 36.

\bibitem[{Lang et~al.(2022)Lang, Vijayaraghavan, and Sontag}]{lang2022training}
Hunter Lang, Aravindan Vijayaraghavan, and David Sontag. 2022.
\newblock Training subset selection for weak supervision.
\newblock \emph{Advances in Neural Information Processing Systems}, 35:16023--16036.

\bibitem[{Li et~al.(2021)Li, Lin, Ren, Li, Zhou, and Sun}]{li2021dynamic}
Lei Li, Yankai Lin, Shuhuai Ren, Peng Li, Jie Zhou, and Xu~Sun. 2021.
\newblock Dynamic knowledge distillation for pre-trained language models.
\newblock In \emph{Proceedings of the 2021 Conference on Empirical Methods in Natural Language Processing}, pages 379--389.

\bibitem[{Liu(2019)}]{liu2019roberta}
Yinhan Liu. 2019.
\newblock Roberta: A robustly optimized bert pretraining approach.
\newblock \emph{arXiv preprint arXiv:1907.11692}.

\bibitem[{Mishra and Sundaram(2021)}]{mishra2021confidence}
Sourav Mishra and Suresh Sundaram. 2021.
\newblock Confidence conditioned knowledge distillation.
\newblock \emph{arXiv preprint arXiv:2107.06993}.

\bibitem[{Saravia et~al.(2018)Saravia, Liu, Huang, Wu, and Chen}]{saravia-etal-2018-carer}
Elvis Saravia, Hsien-Chi~Toby Liu, Yen-Hao Huang, Junlin Wu, and Yi-Shin Chen. 2018.
\newblock \href {https://doi.org/10.18653/v1/D18-1404} {{CARER}: Contextualized affect representations for emotion recognition}.
\newblock In \emph{Proceedings of the 2018 Conference on Empirical Methods in Natural Language Processing}, pages 3687--3697, Brussels, Belgium. Association for Computational Linguistics.

\bibitem[{Schopf et~al.(2022)Schopf, Braun, and Matthes}]{schopf2022evaluating}
Tim Schopf, Daniel Braun, and Florian Matthes. 2022.
\newblock Evaluating unsupervised text classification: zero-shot and similarity-based approaches.
\newblock In \emph{Proceedings of the 2022 6th International Conference on Natural Language Processing and Information Retrieval}, pages 6--15.

\bibitem[{Shoeybi et~al.(2019)Shoeybi, Patwary, Puri, LeGresley, Casper, and Catanzaro}]{shoeybi2019megatron}
Mohammad Shoeybi, Mostofa Patwary, Raul Puri, Patrick LeGresley, Jared Casper, and Bryan Catanzaro. 2019.
\newblock Megatron-lm: Training multi-billion parameter language models using model parallelism.
\newblock \emph{arXiv preprint arXiv:1909.08053}.

\bibitem[{Sohn et~al.(2020)Sohn, Berthelot, Carlini, Zhang, Zhang, Raffel, Cubuk, Kurakin, and Li}]{sohn2020fixmatch}
Kihyuk Sohn, David Berthelot, Nicholas Carlini, Zizhao Zhang, Han Zhang, Colin~A Raffel, Ekin~Dogus Cubuk, Alexey Kurakin, and Chun-Liang Li. 2020.
\newblock Fixmatch: Simplifying semi-supervised learning with consistency and confidence.
\newblock \emph{Advances in neural information processing systems}, 33:596--608.

\bibitem[{Team et~al.(2024)Team, Mesnard, Hardin, Dadashi, Bhupatiraju, Pathak, Sifre, Rivi{\`e}re, Kale, Love et~al.}]{team2024gemma}
Gemma Team, Thomas Mesnard, Cassidy Hardin, Robert Dadashi, Surya Bhupatiraju, Shreya Pathak, Laurent Sifre, Morgane Rivi{\`e}re, Mihir~Sanjay Kale, Juliette Love, et~al. 2024.
\newblock Gemma: Open models based on gemini research and technology.
\newblock \emph{arXiv preprint arXiv:2403.08295}.

\bibitem[{Thirunavukarasu et~al.(2023)Thirunavukarasu, Ting, Elangovan, Gutierrez, Tan, and Ting}]{thirunavukarasu2023large}
Arun~James Thirunavukarasu, Darren Shu~Jeng Ting, Kabilan Elangovan, Laura Gutierrez, Ting~Fang Tan, and Daniel Shu~Wei Ting. 2023.
\newblock Large language models in medicine.
\newblock \emph{Nature medicine}, 29(8):1930--1940.

\bibitem[{Touvron et~al.(2023)Touvron, Lavril, Izacard, Martinet, Lachaux, Lacroix, Rozi{\`e}re, Goyal, Hambro, Azhar et~al.}]{touvron2023llama}
Hugo Touvron, Thibaut Lavril, Gautier Izacard, Xavier Martinet, Marie-Anne Lachaux, Timoth{\'e}e Lacroix, Baptiste Rozi{\`e}re, Naman Goyal, Eric Hambro, Faisal Azhar, et~al. 2023.
\newblock Llama: Open and efficient foundation language models.
\newblock \emph{arXiv preprint arXiv:2302.13971}.

\bibitem[{Wang et~al.(2024)Wang, Zhang, Zhang, Wu, Mo, Lu, Wang, Li, Xu, Tang et~al.}]{wang2024comprehensive}
Fali Wang, Zhiwei Zhang, Xianren Zhang, Zongyu Wu, Tzuhao Mo, Qiuhao Lu, Wanjing Wang, Rui Li, Junjie Xu, Xianfeng Tang, et~al. 2024.
\newblock A comprehensive survey of small language models in the era of large language models: Techniques, enhancements, applications, collaboration with llms, and trustworthiness.
\newblock \emph{arXiv preprint arXiv:2411.03350}.

\bibitem[{Wang et~al.(2023)Wang, Chen, Heng, Hou, Fan, Wu, Wang, Savvides, Shinozaki, Raj et~al.}]{freematch}
Yidong Wang, Hao Chen, Qiang Heng, Wenxin Hou, Yue Fan, Zhen Wu, Jindong Wang, Marios Savvides, Takahiro Shinozaki, Bhiksha Raj, et~al. 2023.
\newblock Freematch: Self-adaptive thresholding for semi-supervised learning.
\newblock In \emph{Eleventh International Conference on Learning Representations}. OpenReview. net.

\bibitem[{Xu et~al.(2023)Xu, Liu, and Loy}]{xu2023computation}
Guodong Xu, Ziwei Liu, and Chen~Change Loy. 2023.
\newblock Computation-efficient knowledge distillation via uncertainty-aware mixup.
\newblock \emph{Pattern Recognition}, 138:109338.

\bibitem[{Yin et~al.(2019)Yin, Hay, and Roth}]{yin2019benchmarking}
Wenpeng Yin, Jamaal Hay, and Dan Roth. 2019.
\newblock Benchmarking zero-shot text classification: Datasets, evaluation and entailment approach.
\newblock In \emph{Proceedings of the 2019 Conference on Empirical Methods in Natural Language Processing and the 9th International Joint Conference on Natural Language Processing (EMNLP-IJCNLP)}, pages 3914--3923.

\bibitem[{Yu et~al.(2023)Yu, Zhang, Xu, Zhang, Shen, and Zhang}]{yu2023cold}
Yue Yu, Rongzhi Zhang, Ran Xu, Jieyu Zhang, Jiaming Shen, and Chao Zhang. 2023.
\newblock Cold-start data selection for better few-shot language model fine-tuning: A prompt-based uncertainty propagation approach.
\newblock In \emph{Proceedings of the 61st Annual Meeting of the Association for Computational Linguistics (Volume 1: Long Papers)}, pages 2499--2521.

\bibitem[{Zhang et~al.(2021)Zhang, Wang, Hou, Wu, Wang, Okumura, and Shinozaki}]{zhang2021flexmatch}
Bowen Zhang, Yidong Wang, Wenxin Hou, Hao Wu, Jindong Wang, Manabu Okumura, and Takahiro Shinozaki. 2021.
\newblock Flexmatch: Boosting semi-supervised learning with curriculum pseudo labeling.
\newblock \emph{Advances in Neural Information Processing Systems}, 34:18408--18419.

\bibitem[{Zhang et~al.(2015)Zhang, Zhao, and LeCun}]{zhang2015character}
Xiang Zhang, Junbo Zhao, and Yann LeCun. 2015.
\newblock Character-level convolutional networks for text classification.
\newblock \emph{Advances in neural information processing systems}, 28.

\bibitem[{Zhao et~al.(2023)Zhao, Zhou, Li, Tang, Wang, Hou, Min, Zhang, Zhang, Dong et~al.}]{zhao2023survey}
Wayne~Xin Zhao, Kun Zhou, Junyi Li, Tianyi Tang, Xiaolei Wang, Yupeng Hou, Yingqian Min, Beichen Zhang, Junjie Zhang, Zican Dong, et~al. 2023.
\newblock A survey of large language models.
\newblock \emph{arXiv preprint arXiv:2303.18223}.

\bibitem[{Zhou et~al.(2023)Zhou, Li, Liu, Guan, Xing, Chen, and Sun}]{zhou2023adads}
Qinhong Zhou, Peng Li, Yang Liu, Yuyang Guan, Qizhou Xing, Ming Chen, and Maosong Sun. 2023.
\newblock Adads: Adaptive data selection for accelerating pre-trained language model knowledge distillation.
\newblock \emph{AI Open}, 4.

\end{thebibliography}

\clearpage
\appendix

\section{Appendices}
\label{sec:appendix}

\begin{table}[h]
\caption {Teacher prompt to generate pseudo-labels. }
\label{table:prompt_teacher}
\centering
\rule{0.48\textwidth}{1.5pt}
\parbox{\textwidth}{
\vspace{5pt} 
<system prompt> \\
Examples: <a few examples from validation set> \\
Sentence: <input text>\\
Answer is: <label> \\
}
\vspace{2pt}
\rule{0.48\textwidth}{1.5pt}
\vspace{5pt}
\end{table}

\subsection{ Prompts }
\label{sec:app_prompt}
We provide more details about the prompts used for the teacher and student models. The teacher prompt template is shown in Table~\ref{table:prompt_teacher} and consists of several components: a system prompt with task instructions, examples from the validation set demonstrating the task, the input text, and the output label. The system prompt for each dataset is listed in Table~\ref{table:sys_prompt}. 
To determine the number of few-shot examples, we tried number of 3, 5, 10 and chose the one with the best validation result. Specifically, we set the few-shot number  as follows:   PubMed-RCT-20k (5), Yahoo! Answers (5), Emotions (5), Arxiv-10 (3), and BiosBias (10). 
For the student model, we use the template ``<Input text>. It is [MASK]".

\subsection{Additional Results of Empirical Observations}
\label{sec:app_emp}
\begin{figure*}[t]
\begin{center}
 \centerline{
\subfigure[Yahoo! Answers]{
\includegraphics[width=0.24\linewidth]{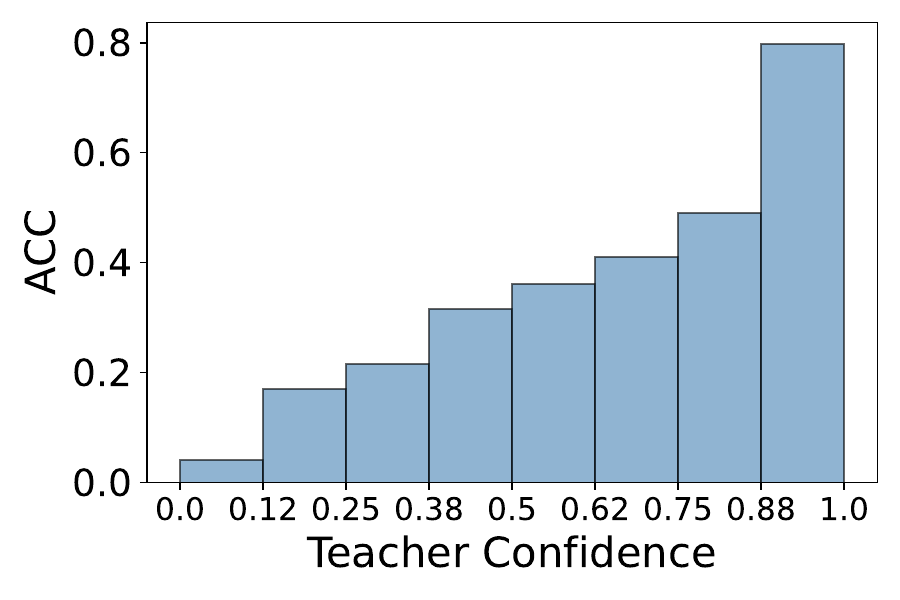} 
}

\subfigure[Emotions]{
\includegraphics[width=0.24\linewidth]{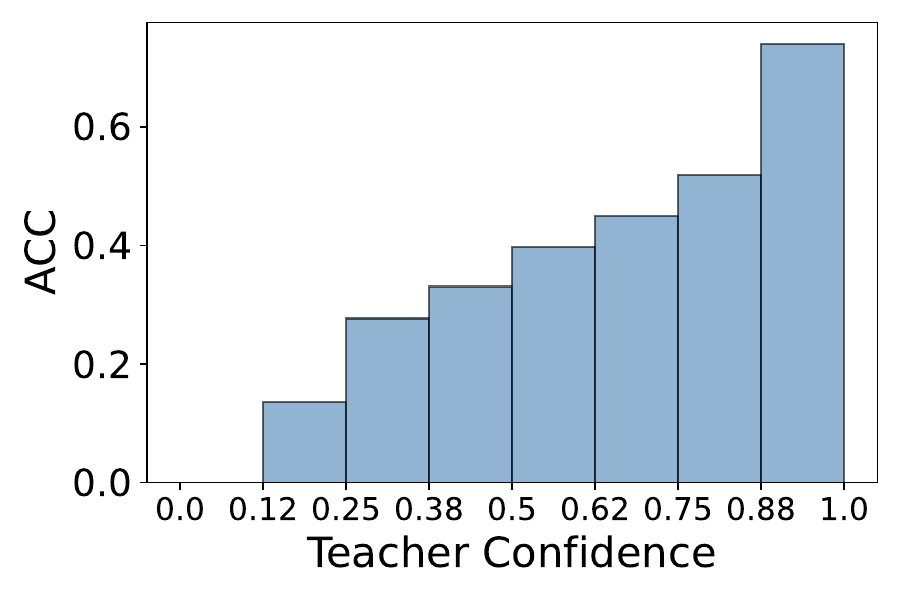} 
}
\subfigure[Arxiv-10]{
\includegraphics[width=0.24\linewidth]{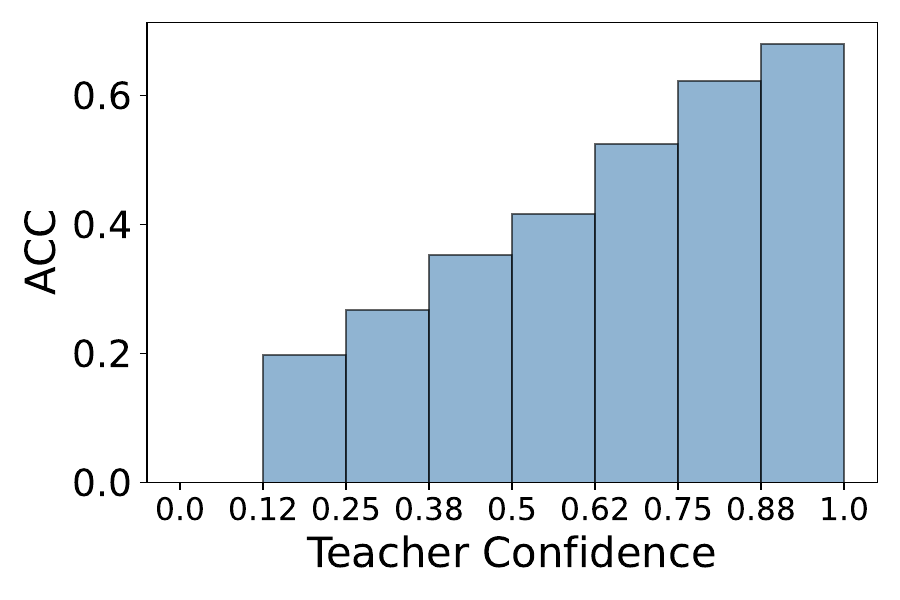} 
}
\subfigure[BiosBias]{
\includegraphics[width=0.24\linewidth]{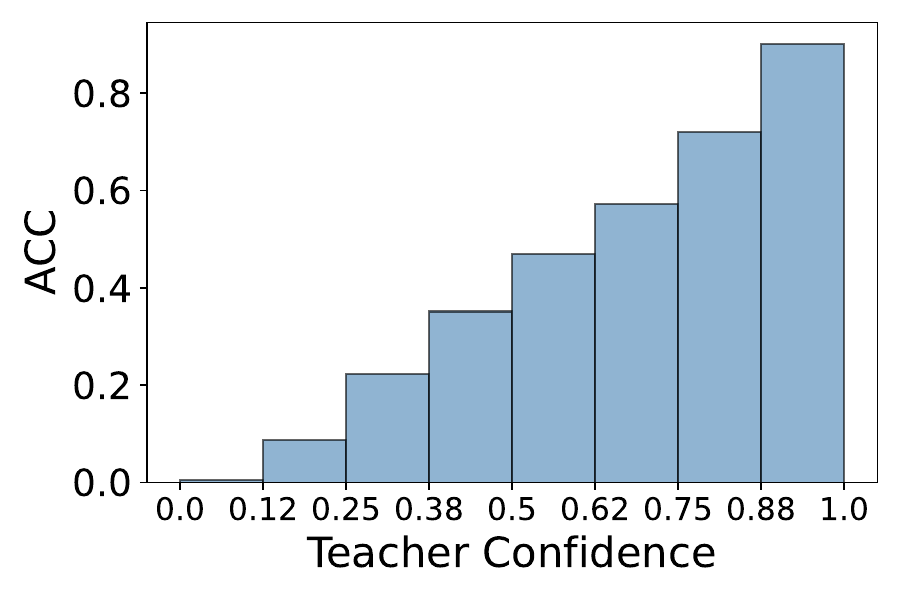} 
}
}

\centerline{
\subfigure[Yahoo! Answers]{
\includegraphics[width=0.24\linewidth]{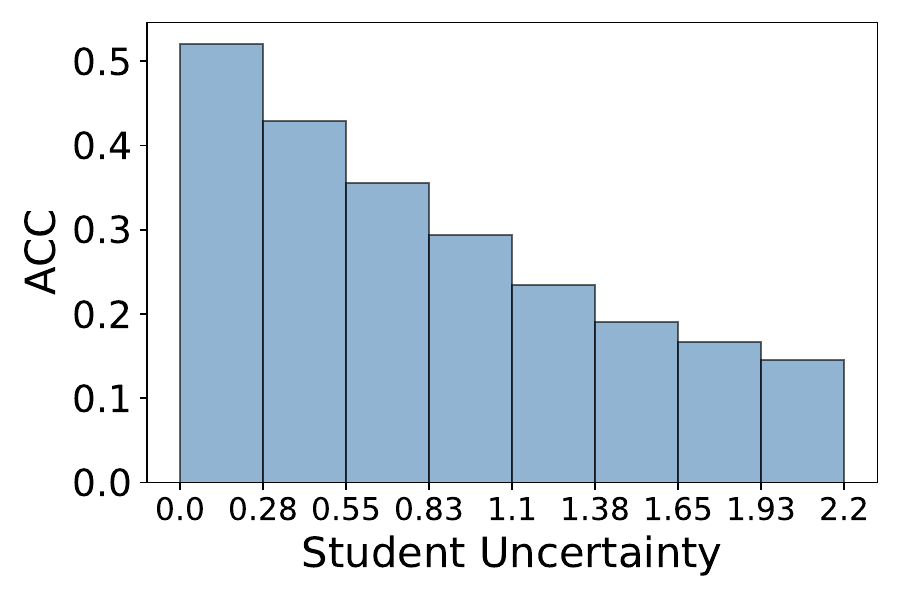} 
}
\subfigure[Emotions]{
\includegraphics[width=0.24\linewidth]{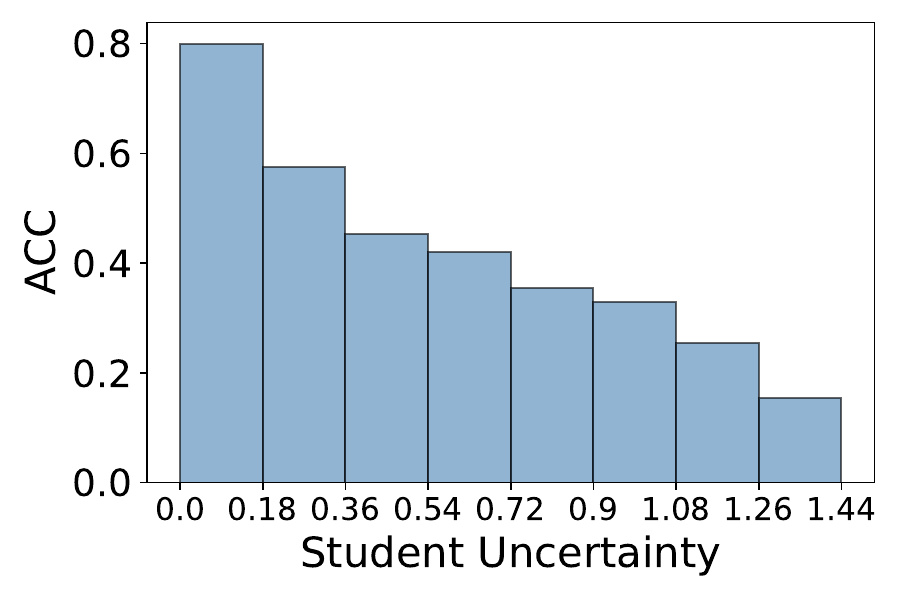} 
}
\subfigure[Arxiv-10]{
\includegraphics[width=0.24\linewidth]{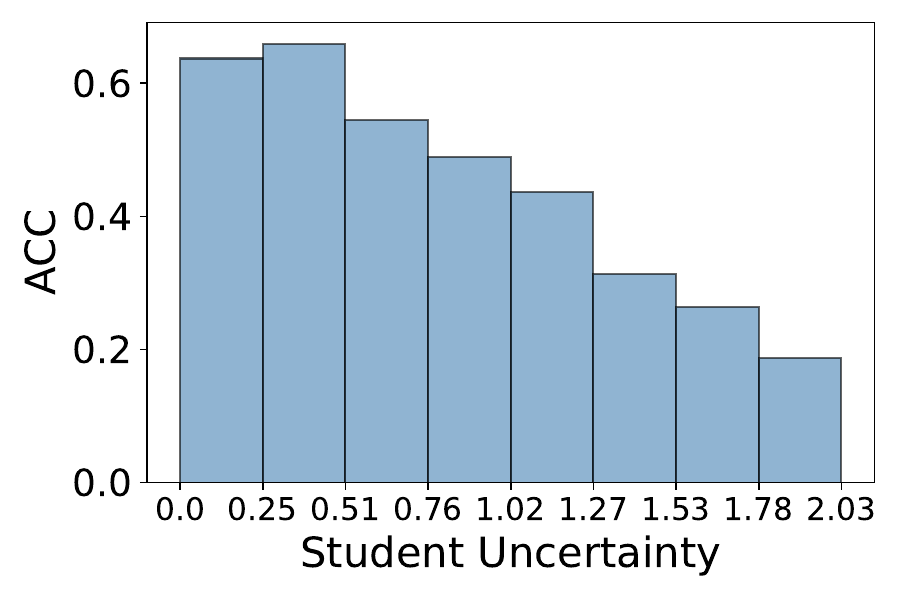} 
}
\subfigure[BiosBias]{
\includegraphics[width=0.24\linewidth]{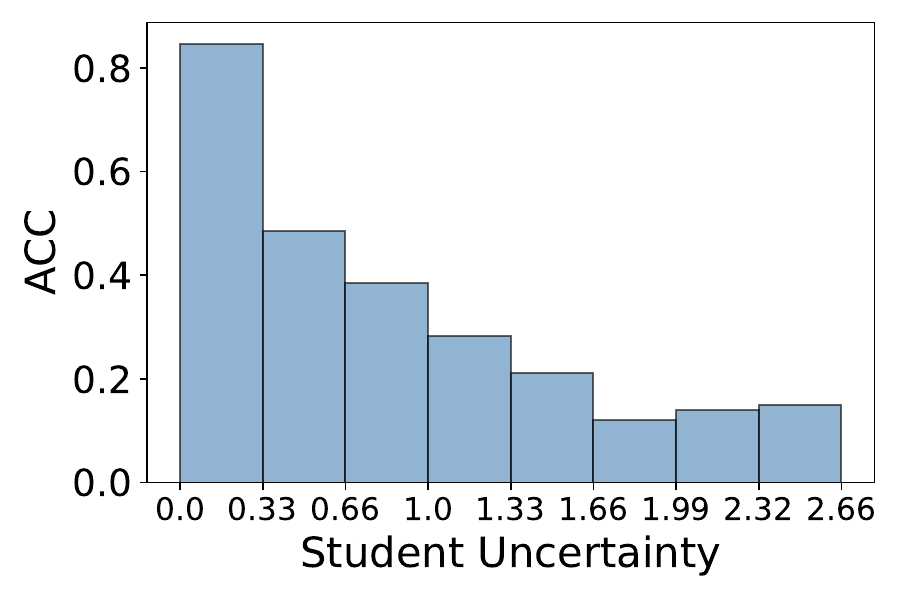} 
}
}
 
\caption{The relationship between teacher model accuracy and teacher confidence (a)-(d), and the relationship between student model accuracy and student uncertainty (e)-(h) on the validation set of various dataset.    }
\label{fig:Acc_teacher_student_app}
\end{center}
\vspace{-0.2in}
\end{figure*}

In this subsection, we explore the relationship between teacher model accuracy and teacher confidence, as well as the relationship between student model accuracy and student uncertainty across additional datasets. The results are  shown in \figurename~\ref{fig:Acc_teacher_student_app}. They reveal similar trends to those observed with the PubMed-RCT-20k dataset: higher teacher confidence generally correlates with higher teacher accuracy, and greater student uncertainty typically corresponds to incorrect predictions, i.e., hard samples. These findings further validate that teacher confidence can be used to assess pseudo-label quality, while student uncertainty can be used to access the informativeness of samples.

\begin{table}[!h]
\centering

 \caption{Dataset Licenses.}
  \begin{adjustbox}{width =0.45 \textwidth}
\begin{tabular}{c|c}
\toprule
 Datasets	&License	 \\
  \midrule
 PubMed-RCT-20k & No License \\
 Yahoo! Answers & CC-BY-SA\\
 Emotions & CC-BY-SA\\
 Arxiv-10 & GNU General Public License \\
 BiosBias & MIT License\\
\bottomrule
\end{tabular}
\label{table:app_license}
\end{adjustbox}
\end{table}

\subsection{Datasets}
\label{sec:app_data}
In this subsection, we provide more details about the datasets used in our experiments, which are summarized in Table~\ref{table:data}. All datasets are in English, and their respective licenses are listed in Table~\ref{table:app_license}. 

\begin{itemize}
    \item 

\textbf{PubMed-RCT-20k}\footnote{\url{https://github.com/Franck-Dernoncourt/pubmed-rct?tab=readme-ov-file}}\citep{dernoncourt2017pubmed}  is a sequential sentence classification dataset derived from PubMed abstracts, where each sentence is labeled with roles such as background, objective, method, result, or conclusion. The dataset contains 20k abstracts and has fixed train, validation and test splits.

\item \textbf{Yahoo! Answers}\footnote{\url{https://github.com/LC-John/Yahoo-Answers-Topic-Classification-Dataset/tree/master}}~\citep{zhang2015character} 
 is a topic classification dataset collected from the Yahoo Answers platform consisting of 10 classes. Due to computational limitations, we randomly select a subset: 10k samples from each class for training, and 6k samples from each class for validation and test.

\item \textbf{Emotions}\footnote{\url{https://huggingface.co/datasets/dair-ai/emotion}}~\citep{saravia-etal-2018-carer} is a dataset of Twitter messages categorized into six basic emotions: anger, fear, joy, love, sadness, and surprise. Since the dataset does not have fixed splits, we randomly split it into train, validation, and test sets using an 80\%, 10\%, 10\% ratio.

\item \textbf{Arxiv-10}\footnote{\url{https://paperswithcode.com/dataset/arxiv-10}}~\citep{farhangi2022protoformer} 
consists of abstracts and titles from 100,000 ArXiv papers, balanced across 10 classes in fields such as computer science, physics, and mathematics. This dataset also lacks fixed splits, so we randomly split it into train, validation, and test sets using an 80\%, 10\%, 10\% ratio.

\item \textbf{BiosBias}\footnote{\url{https://huggingface.co/datasets/LabHC/bias_in_bios?row=46}}~\citep{De-ArteagaRWCBC19} 
 contains textual biographies aimed at predicting professional occupations with 28 classes. This dataset includes predefined splits for training, validation and testing.
\end{itemize}

By following current works~\citep{kontonis2024slam, iliopoulos2022weighted}, we randomly select 500 labeled samples  as the  validation set from the original validation set.  

\begin{table}[!ht]
\centering
\caption{ ACC and Macro-F1  (\%) of the  text classification. Average results with standard deviation based on 3 seeds are reported.
}
\label{table:true_label}
\begin{adjustbox}{width =0.5 \textwidth}
\begin{tabular}{l|cc|cc}
\toprule
Models &  \multicolumn{2}{c|}{Emotions} & \multicolumn{2}{c}{Arxiv-10}  \\
\midrule
& ACC & F1 &ACC &F1\\
\midrule
True Label &47.93 ± 2.32	& 23.10 ± 5.24 &   53.33 ± 2.72	&53.82 ± 1.80\\
LLKD & \textbf{67.04 ± 1.12}	& \textbf{55.42 ± 2.62 }&	 \textbf{62.90 ± 0.99}&	  \textbf{62.18 ± 1.08}\\
\bottomrule
\end{tabular}
\end{adjustbox}
\vspace{-0.1in}
\end{table}

\begin{table*}[!ht]
\centering
\caption{ Data statistics. \#Train, \#Val, \#Test, and \#Class are the number of training, validation, test and class respectively.}
\label{table:data}
\begin{adjustbox}{width =1 \textwidth}
\begin{tabular}{lccccl}
\toprule
Datasets & \#Train & \#Val & \#Test &\#Class&Labels \\
\midrule
PubMed-RCT-20k & 180,030 & 500 & 30,135& 5  & BACKGROUND, OBJECTIVE, METHODS, RESULTS, CONCLUSIONS\\
Yahoo! Answers & 99,970 &500&60,000&10 & \makecell[l]{society, science, health, education, computer, sports, business, entertainment, \\ relationship, politics} \\
Emotions & 333,431&500 &41,680 & 6 & sadness, joy, love, anger, fear, surprise\\

 Arxiv-10 &79,970 &500 & 10,000 & 10& astro-ph, cond-mat, cs, eess, hep-ph, hep-th, math, physics, quant-ph, stat\\
 BiosBias & 257,394& 500 & 99,100 & 28& \makecell[l]{accountant,architect, attorney, doctor, comedian, composer, dentist, \\ nutritionist,  dj, filmmaker, designer,journalist, model, nurse, painter, \\assistant, 
pastor, trainer, photographer, physician, poet, professor, \\psychologist, rapper, programmer, surgeon,teacher, instructor}\\

\bottomrule
\end{tabular}
\end{adjustbox}
\end{table*}

\subsection{Baselines and Implementaion Details}
\label{sec:app_settings}

All baseline models, except for the teacher and Lbl2TransformerVec, use RoBERTa as the backbone model and share the same pseudo-labels as our model. For the teacher model, we use Llama-3-8B-Instruct\footnote{\url{https://huggingface.co/meta-llama/Meta-Llama-3-8B-Instruct}}, while RoBERTa-base\footnote{\url{https://huggingface.co/FacebookAI/roberta-base}} serves as the student model.
We tune the learning rate within the range of {1e-4, 1e-5}, and $\lambda_T$ and $\lambda_S$ in the range of \{0.1, 0.3, 0.5, 0.7, 0.9\}. The values of $\beta_{S1}$, $\beta_{S2}$, $\beta_{T1}$, and $\beta_{T2}$ are set in \{0, 1\}. The batch size is set to 32. The implementation is based on two open-source packages: VLLM\footnote{\url{https://docs.vllm.ai/en/latest/}} and OpenPrompt\footnote{\url{https://github.com/thunlp/OpenPrompt}}.
The experiments were conducted using an NVIDIA-A10 with 23GB of memory and an NVIDIA-A100 with 40GB.

\subsection{Exploration of Using True Label}
\label{sec:app_true_label}
 When the validation set has true labels and is limited to a small size (e.g., 500 samples), one straightforward approach is to just using these labeled samples into the student model’s training. However, we observed that such a small labeled set is typically insufficient for effective student training. To further investigate, we divided the 500 validation samples into two parts: 50 samples for validation and the remaining 450 samples for training. This setting is denoted as "True Label" in the Table~\ref{table:true_label}, with results shown for the Emotions and Arxiv-10 datasets.
Our observations indicate that this setting performs poorly, with significantly worse results than the baselines and a substantial gap compared to our proposed LLKD model. These findings emphasize the necessity of leveraging unlabeled data when the labeled dataset is limited.

\subsection{Additional Results of Thresholds Evaluation}
\label{sec:app_threshold}

\begin{figure*}[t]
\begin{center}
 \centerline
{

\subfigure[PubMed-RCT-20k]{
\includegraphics[width=0.5\linewidth]{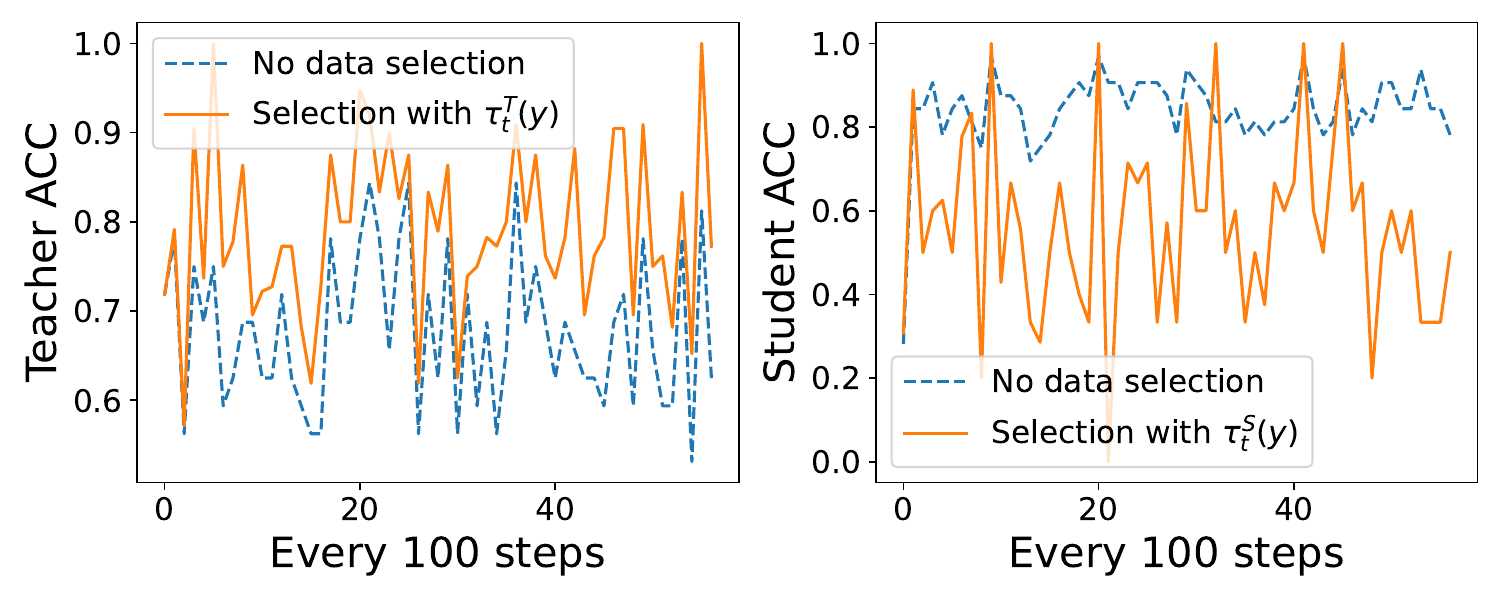} 
}

\subfigure[Emotions]{
\includegraphics[width=0.5\linewidth]{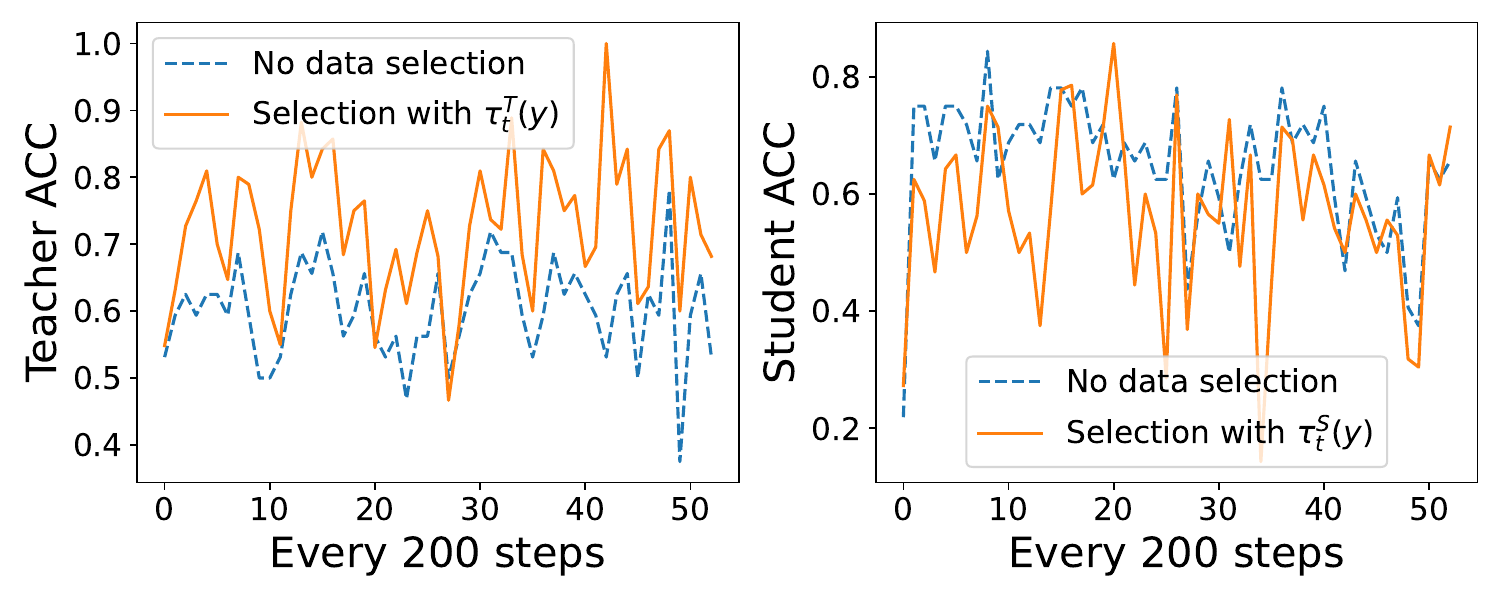} 
}

}
\centerline{
\subfigure[Yahoo! Answers]{
\includegraphics[width=0.5\linewidth]{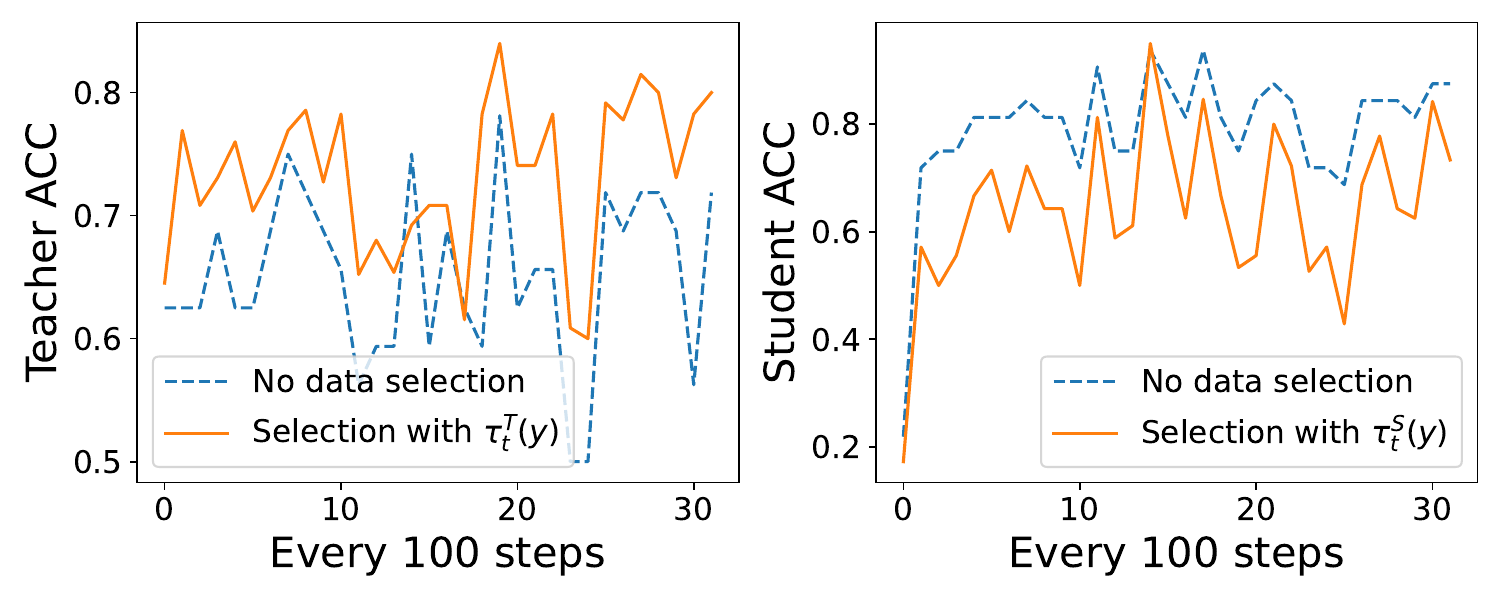} 
}
\subfigure[BiosBias]{
\includegraphics[width=0.5\linewidth]{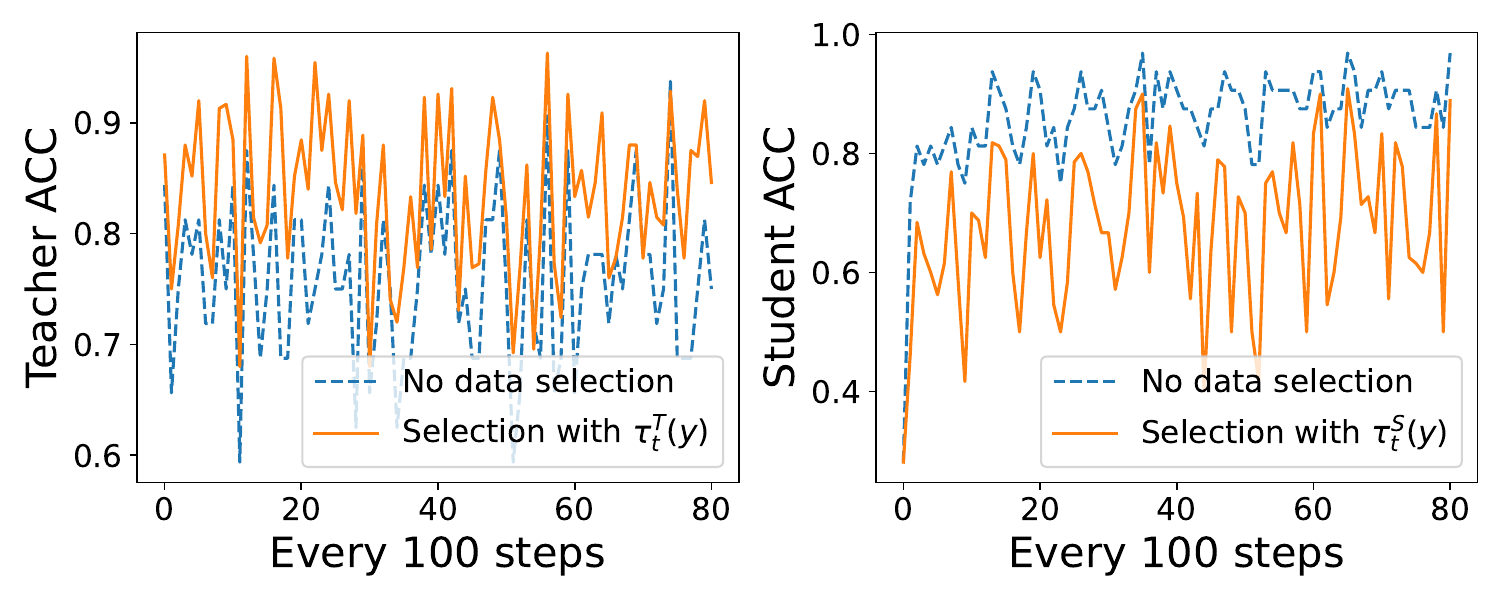} 
}
 }

\caption{Teacher ACC and student ACC before and after data selection on various datasets.  }
\label{fig:app_threshold}
\end{center}

\end{figure*}

In this subsection, we present additional results from the thresholds evaluation on other datasets, as shown in \figurename~\ref{fig:app_threshold}. We observe similar trends to those seen in the Arxiv-10 dataset, confirming that selecting samples with values larger than $\tau_t^T(y)$ yields high-quality pseudo-labels, while larger $\tau_t^S(y)$ selects more challenging samples.

\begin{table*}
\centering
\caption{The system prompts on the teacher prompt for each dataset.}
\mycfs{9.5}

\begin{tabular}{p{2.5cm}p{12.6cm}}
\toprule
\textbf{Dataset} & \textbf{System Prompt}  \\
\midrule
{PubMed-RCT-20k} & Assign the correct category to the sentences below based on the provided examples. The sentences are from scientific abstracts and belong to one of the following classes: BACKGROUND (A), OBJECTIVE (B), METHODS (C), RESULTS (D), or CONCLUSIONS (E). BACKGROUND sentences provide necessary context. OBJECTIVE sentences state the research goal. METHODS sentences describe the methodology. RESULTS sentences present the findings. CONCLUSIONS sentences discuss the implications. Indicate the class using A/B/C/D/E.  \\
\midrule

Yahoo! Answers & Determine the category for each of the following sentences based on the context provided. The categories include: society, science, health, education, computers, sports, business, entertainment, relationship, and politics. society relates to social and cultural contexts; science focuses on empirical and theoretical knowledge; health involves wellness and medical discussions; education pertains to academic matters; computers is about technology and IT; sports involves physical activities; business covers financial and commercial activities; entertainment deals with media and arts; relationship is about human connections; politics covers governance and public policy. \\
\midrule
Emotions & Classify the given Twitter message into one of the following emotions: sadness, joy, love, anger, fear, or surprise. Sadness reflects a deep emotional state of sorrow or disappointment. Joy conveys happiness and positive feelings. Love indicates a strong emotional bond. Anger represents irritation or annoyance towards something. Fear is related to feelings of anxiety or worry. Surprise is a reaction to something unforeseen or unexpected.\\
\midrule
Arxiv-10 & Analyze the following sentences and determine the appropriate category. The possible categories are: A for astro-ph (astronomy), B for cond-mat (condensed matter physics), C for cs (computer science), D for eess (electrical engineering and system sciences), E for hep-ph (high-energy physics phenomenology), F for hep-th (theoretical high-energy physics), G for math (mathematics), H for physics (general physics), I for quant-ph (quantum mechanics), and J for stat (statistics). Choose the correct label using A/B/C/D/E/F/G/H/I/J. \\
\midrule
BiosBias & Given the sentences below, identify the correct occupation from a list of 28 options. These options include but are not limited to: accountant, architect, attorney, chiropractor, comedian, composer, dentist, dietitian, dj, filmmaker, interior designer, journalist, model, nurse, painter, paralegal, pastor, personal trainer, photographer, physician, poet, professor, psychologist, rapper, software engineer, surgeon, teacher, yoga teacher\\
\bottomrule
\end{tabular}
\vspace{-0.2cm}
\label{table:sys_prompt}
\end{table*}


%


%

\end{document}